\documentclass[10pt, conference, compsocconf]{IEEEtran}
\usepackage{xcolor}
\usepackage{mathtools}
\usepackage{enumerate}
\usepackage{amssymb}
\usepackage{amsmath}
\usepackage{eqnarray}
\usepackage[]{algorithm}
\usepackage{clrscode3e}
\usepackage[pdftex]{graphicx}

\usepackage[font={small}]{caption,subfig}  
\DeclarePairedDelimiter{\ceil}{\lceil}{\rceil}

\begin{document}


\title{ZNN -- A Fast and Scalable Algorithm for Training 3D
  Convolutional Networks on Multi-Core and Many-Core Shared Memory
  Machines}

\author{\IEEEauthorblockN{Aleksandar Zlateski\IEEEauthorrefmark{1},
    Kisuk Lee\IEEEauthorrefmark{2}}
  \IEEEauthorblockA{\IEEEauthorrefmark{1}Electrical Engineering and
    Computer Science Dept.\\ \IEEEauthorrefmark{2}Brain and
    Cognitive Sciences Dept.\\ Massachusetts Institute of
    Technology\\ Cambridge, MA 02139 USA\\ \IEEEauthorrefmark{1}{\tt
      zlateski@mit.edu}, \IEEEauthorrefmark{2}{\tt kisuklee@mit.edu}}
  \and \IEEEauthorblockN{H. Sebastian Seung}
  \IEEEauthorblockA{Princeton Neuroscience Institute\\ and Computer
    Science Dept.\\ Princeton University\\ Princeton, NJ 08540
    USA\\ {\tt sseung@princeton.edu} }}


\maketitle


\begin{abstract}
Convolutional networks (ConvNets) have become a popular approach to
computer vision. It is important to accelerate ConvNet training, which
is computationally costly.  We propose a novel parallel algorithm
based on decomposition into a set of tasks, most of which are
convolutions or FFTs. Applying Brent's theorem to the task dependency
graph implies that linear speedup with the number of processors is
attainable within the PRAM model of parallel computation, for wide
network architectures. To attain such performance on real
shared-memory machines, our algorithm computes convolutions converging
on the same node of the network with temporal locality to reduce cache
misses, and sums the convergent convolution outputs via an almost
wait-free concurrent method to reduce time spent in critical sections.
We implement the algorithm with a publicly available software package
called ZNN.  Benchmarking with multi-core CPUs shows that ZNN can
attain speedup roughly equal to the number of physical cores.  We also
show that ZNN can attain over 90x speedup on a many-core CPU (Xeon
Phi\texttrademark Knights Corner).  These speedups are achieved for
network architectures with widths that are in common use.  The task
parallelism of the ZNN algorithm is suited to CPUs, while the SIMD
parallelism of previous algorithms is compatible with GPUs.  Through
examples, we show that ZNN can be either faster or slower than certain
GPU implementations depending on specifics of the network
architecture, kernel sizes, and density and size of the output patch.
ZNN may be less costly to develop and maintain, due to the relative
ease of general-purpose CPU programming.

\end{abstract}


\section{Introduction}
A standard formulation of supervised learning starts with a
parametrized class of mappings, a training set of desired input-output
pairs, and a loss function measuring deviation of actual output from
desired output.  The goal of learning is to minimize the average loss
over the training set. A popular minimization method is stochastic
gradient descent. For each input in sequence, the parameters of the
mapping are updated in minus the direction of the gradient of the loss
with respect to the parameters.  Here we are concerned with a class of
mappings known as convolutional networks (ConvNets).

Significant effort has been put into parallelizing ConvNet learning on
GPUs, as in the popular software packages Caffe~\cite{jia2014caffe},
Torch~\cite{collobert2011torch7} and Theano\cite{bergstra2010theano}.
ConvNet learning has also been distributed over multiple machines
~\cite{dean2012large}.  However, there has been relatively little
work on parallelizing ConvNet learning for single shared memory CPU
machines.

Here we introduce a software package called ZNN, which implements a
novel parallel algorithm for ConvNet learning on multi-core and
many-core CPU machines.  ZNN implements 3D ConvNets, with 2D as a
special case.  ZNN can employ either direct or FFT convolution, and
chooses between the two methods by autotuning each layer of the
network.  FFT convolution was previously applied to 2D ConvNets
running on GPUs~\cite{mathieu-iclr-14,vasilache2014fast}, and is
even more advantageous for 3D ConvNets on CPUs.

As far as we know, ZNN is the first publicly available software that
supports efficient training of sliding window max-pooling ConvNets,
which have been studied
by~\cite{masci2013fast,giusti2013fast,sermanet2013overfeat}.

There is related work on using Xeon Phi\texttrademark for supervised
deep learning ~\cite{viebke2015potential}.  and unsupervised deep
learning~\cite{Jin:2014:TLS:2672598.2672903}.

\section{Computation graph}

\begin{figure}
  \begin{center}
  \includegraphics[width=0.8\columnwidth]{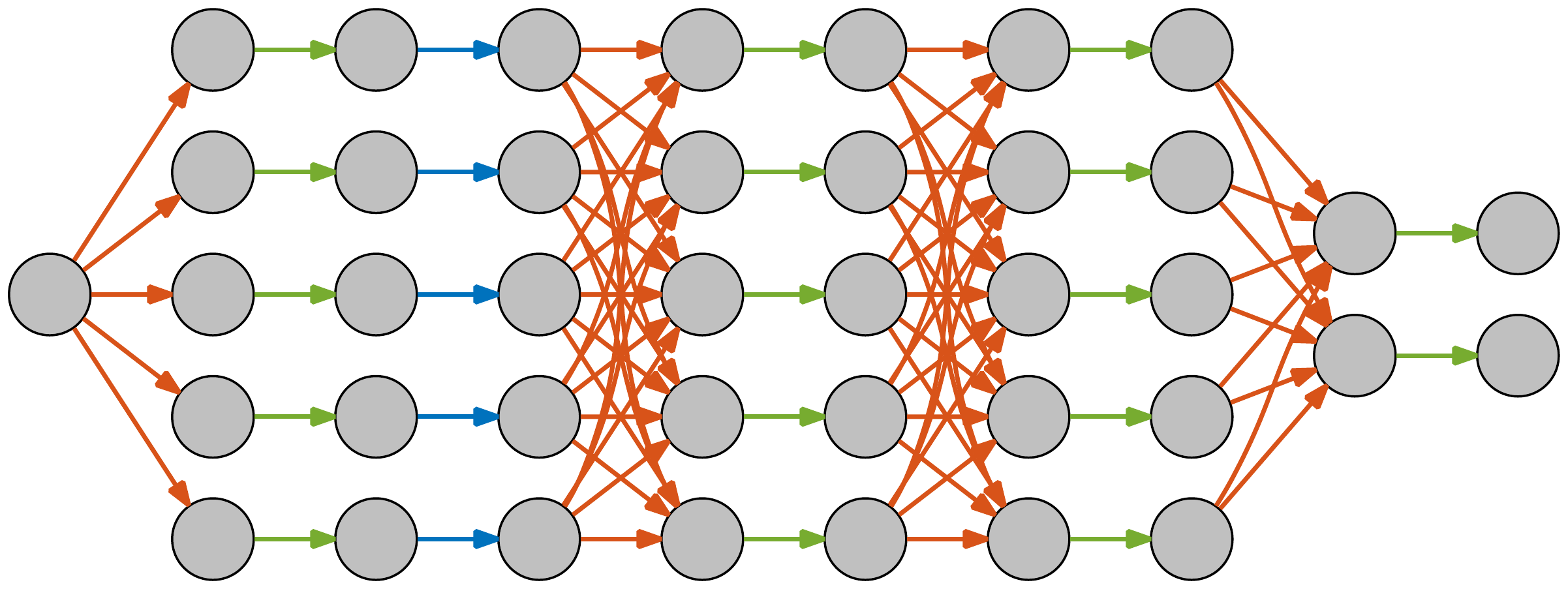}
  \end{center}
  \caption{ConvNet computation graph.  Leftmost node is input image,
    and rightmost nodes are output images. Edges represent convolution
    (red), transfer function (green), and max pooling/filtering (blue)
    operations.}
  \label{fig:convnet_dag}
\end{figure}

We define a ConvNet using a directed acyclic graph (DAG), called the
\emph{computation graph} (Fig. \ref{fig:convnet_dag}).  Each node
represents a 3D image, and each edge some image filtering operation.
(2D images are a special case in which one of the dimensions has size
one.)  If multiple edges converge on a node, the node sums the outputs
of the filtering operations represented by the edges. For convenience,
the discussion below will assume that images and kernels have
isotropic dimensions, though this restriction is not necessary for
ZNN.  The image filtering operations are of the four following types.

{\bf Convolution} A weighted linear combination of voxels within a
sliding window is computed for each location of the window in the
image.  The set of weights of the linear combination is called the
\emph{kernel}.  If the input image has size $n^3$ and the kernel has
size $k^3$, then the output image has size $n'^3 = (n-k+1)^3$. Image
size decreases because an output voxel only exists when the sliding
window is fully contained in the input image.\footnote{This is known
  as a \texttt{valid} convolution in MATLAB.}
The convolution is allowed to be sparse, meaning that only every $s$th
image voxel (in every dimension) within the sliding window enters the
linear combination.

{\bf Max-pooling} divides an image of size $n^3$ into blocks of size
$p^3$, where $n$ is divisible by $p$. The maximum value is computed
for each block, yielding an image of size $(n/p)^3$.

{\bf Max-filtering} The maximum within a sliding window is computed
for each location of the window in the image.  For a window of size
$k^3$ and an input image of size $n^3$, the output image has size
$(n-k+1)^3$. 3D max-filtering can be performed by sequential 1D
max-filtering of $n^2$ arrays in each of the three directions.  For
each array we keep a heap of size $k$ containing the values inside the
1D sliding window.  Each element of the array will be inserted and
removed at most once, each operation taking $\log k$.  For each
position of the sliding window the top of the heap will contain the
maximum value.

\begin{table}
  \centering
  \begin{tabular}{llll}
    \hline
    Pass    &Pooling   &Filtering    &Transfer function
    \\ \hline
    Forward & $f \cdot n^3$ & $f \cdot 6n^3 \log k$ & $f \cdot n^3$
    \\
    Backward & $f \cdot n^3$ & $f \cdot n^3$ & $f \cdot n^3$
    \\
    Update & $-$ & $-$ & $f \cdot n^3$
    \\ \hline
  \end{tabular}
  \caption{Number of floating point operations (FLOPs) required by a
    layer with $f$ nodes that all perform the same nonlinear filtering
    operation (max-pooling, max-filtering, or transfer function).}
  \label{table:other_complexity}
\end{table}

{\bf Transfer function} adds a number called the bias to each voxel of
the image and then applies a nonlinear function to the result.  The
nonlinear function is typically nondecreasing.  Common choices are the
logistic function, the hyperbolic tangent and half-wave rectification.

The computational complexities of max-pooling, max-filtering, and
transfer function are shown in Table~\ref{table:other_complexity}.

For ConvNets in common use, the computation graph has the following
properties:
\begin{itemize}
\item All convergent edges are convolutions; if a node has a sole
  incoming edge, the edge represents a nonlinear filtering operation.
\item Nodes with convergent edges are not adjacent in the graph, but
  are separated from each other by nonlinear filtering edges.  This is
  a reasonable constraint, because a composition of two convolutions
  can be collapsed into a single convolution, thereby simplifying the
  graph.
\item The graph has a layered organization in which all edges in a
  layer represent operations of the same type.
\end{itemize}
ZNN works for general computation graphs, whether or not they possess
the above properties.

\subsection{Sliding window max-pooling ConvNet}

A max-pooling ConvNet in the context of visual object
recognition~\cite{krizhevsky2012imagenet} is a special case of the
definition given above.  No max-filterings are used.  The size of the
input image (known as the ConvNet field of view) is such that the
convolutions and max-poolings reduce the output image(s) to exactly
one pixel/voxel.  There may be a single output representing whether or
not the input image belongs to a given object class, or a set of $n$
outputs representing membership in one of $n$ object classes.

\begin{figure}
  \centering
  \includegraphics[width=0.48\textwidth]{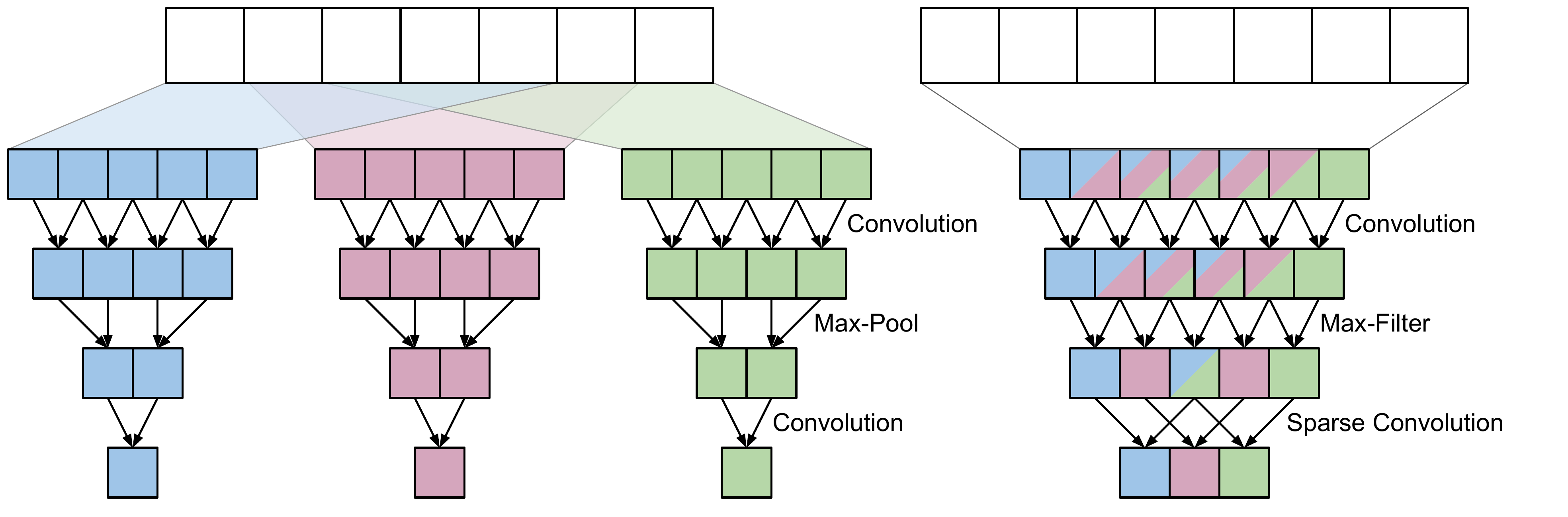}
  \caption{The output of a sliding window max-pooling ConvNet (left)
    can be efficiently computed by a max-filtering ConvNet with sparse
    convolution (right).}
  \label{fig:max_filter}
\end{figure}

If localization and detection are desired as well as recognition, one
can slide a window over a large image, and apply the max-pooling
ConvNet at each location of the window
\cite{sermanet2013overfeat}. For an input image of size $n^3$ and a
ConvNet field of view of size $v^3$, the output image is of size
$(n-v+1)^3$.  The sliding window max-pooling ConvNet is also useful in
the context of boundary detection and image
segmentation~\cite{ciresan2012deep}.  However, it is computationally
wasteful to literally implement the computation in this way.  It is
more efficient to use a max-filtering ConvNet, in which each
max-filtering layer increases the sparsity of all subsequent
convolutions by a factor equal to the size of the max-filtering window
(Fig. \ref{fig:max_filter}).  This approach has been called
\emph{skip-kernels}~\cite{sermanet2013overfeat} or \emph{filter
  rarefaction}~\cite{long2015fully}, and is equivalent in its results
to
\emph{max-fragmentation-pooling}~\cite{giusti2013fast,masci2013fast}.
ZNN can implement the above, but is more general as the sparsity of
convolution need not increase in lock step with max-filtering, but can
be controlled independently.

This sparsity control capability can confer a great deal of
flexibility on ConvNets. It could be useful when implementing a
``scale-invariant'' ConvNet~\cite{kanazawa2014locally}, where
convolutions with shared kernel weights are performed at multiple
scales to capture scale-invariant features.  The scale-invariant
convolution can be easily achieved by controlling the sparsity of
convolutions.  Unlike max-pooling, max-filtering does not decrease the
resolution of filtered images. Thus, every image in max-filtering
ConvNets keeps the original resolution. This is particularly
beneficial to the multi-scale
approach~\cite{long2015fully,sermanet2011traffic}, where images with
multiple resolutions are combined together to construct the
representation. In max-pooling ConvNets, upsampling is commonly used
to adjust the different resolutions of images at multiple
levels. Max-filtering ConvNet, in contrast, removes the need for such
upsampling in an elegant and much more efficient way.

\section{Backpropagation learning}

The backpropagation algorithm is a way of calculating the gradient of
the loss function with respect to the trainable parameters in a
ConvNet, the kernels and biases.  For each input, the calculation
proceeds in several phases:
\begin{enumerate}
\item Obtain an input and desired output from the training set.
\item \emph{Forward pass} - compute the actual output of the ConvNet
  from the input image.
\item Compute the gradient of the loss function with respect to the
  actual output.
\item \emph{Backward pass} - Compute the gradient of the loss function
  with respect to the voxels of the output image at each node.
\item \emph {Weight update} - Compute the gradient of the loss
  function with respect to the kernels and biases, and update these
  parameters in the direction of minus the gradient.
\end{enumerate}
The forward pass has already been described above.  ZNN implements
several possibilities for the loss function, such as the Euclidean
distance between the actual and desired outputs.

\subsection{Backward pass}

It turns out that the backward pass can be represented by another
graph that looks the same as the forward computation graph, except
that the direction of every edge is reversed.  The output nodes of the
forward graph become the input nodes of the backward graph, and are
initialized with the gradient of the loss function with respect to the
voxels of the output nodes of the forward graph.  The nodes of the
backward graph are associated with their own images, which are
distinct from the ones associated with the nodes of the forward graph.

Every edge in the backward graph is multiplication by the transpose of
the Jacobian matrix of the operation represented by the corresponding
edge in the forward computation graph.  The four edge operations in
the forward graph become the following four edge operations in the
backward graph.

{\bf Convolution Jacobian} Convolution in the forward pass becomes
convolution in the backward pass.  The kernel is the same, except that
it is reflected along all three dimensions.
If the input image has size $n^3$ and the kernel has size $k^3$, then
the output image has size $n'^3 = (n+k-1)^3$. Image size increases
because an output voxel exists whenever the sliding window has some
overlap with the input image.\footnote{This is known as a
  \texttt{full} convolution in MATLAB.}

{\bf Max-pooling Jacobian} Within each block, all voxels are zeroed
out except for the one that was identified as the maximum within that
block in the forward pass.  An image of size $n^3$ is expanded into an
image of size $n^3p^3$.

{\bf Max-filtering Jacobian} Every element of an image of size $n'^3 =
(n+p-1)^3$ is initialized to zero.  For each position of the sliding
window the appropriate value of the input is accumulated to the
position from which the maximum element was selected for that window
in the forward pass.

{\bf Transfer function Jacobian} Every voxel of a backward image is
multiplied by the derivative of the transfer function for the
corresponding voxel in the forward image.

\subsection{Weight update}

After the forward and backward passes are complete, there are
``forward images'' at the nodes of the forward computation graph, and
``backward images'' at the nodes of the backward computation graph
(except the input nodes).  These are used to update the kernels and
biases as follows.

{\bf Kernel update} For a convolution going from node $a$ to node $b$
in the forward graph, the gradient of the loss with respect to the
kernel is computed by convolving the reflected forward image at node
$a$ with the backward image at node $b$. A \texttt{valid} convolution
is performed, yielding an image the same size as the kernel.  This is
multiplied by a small ``learning rate parameter,'' and then subtracted
from the kernel.

{\bf Bias update} For a bias at node $a$, the gradient of the loss is
calculated as the sum of all voxels in the backward image at node $a$.
The scalar result is multiplied by a small ``learning rate
parameter,'' and then subtracted from the bias.

\section{Direct vs. FFT convolution}

For a single convolution of an image of size $n^3$ with a kernel of
size $k^3$, it is well-known that the FFT method (complexity
$\mathcal{O}(n^3 \log n)$), becomes more efficient than the direct
method (complexity $\mathcal{O}(n^3k^3)$), for sufficiently large
kernel sizes. The crossover point of equal complexity satisfies
$k^3\sim \log n$.  It is less well-known that the FFT-direct crossover
occurs at smaller kernel sizes for a ConvNet than for a single
convolution~\cite{mathieu-iclr-14,vasilache2014fast}.  This is because
the FFT of an image at a node can be shared by edges at that node (see
Table~\ref{table:conv_complexity}).\footnote{Note that our values
  differ from the ones in~\cite{mathieu-iclr-14} as we take into
  account the difference in complexity between \texttt{full} and
  \texttt{valid} convolutions.}  ZNN performs layerwise auto-tuning to
choose between FFT-based or direct convolution for each laeyer.

Complexity can be further reduced by memoizing the FFTs of images and
kernels obtained during the forward pass for reuse during the backward
pass and weight update.  This possibility was previously noted in
passing but not implemented due to limited onboard GPU memory
~\cite{mathieu-iclr-14,vasilache2014fast}.  The reduction in
complexity is approximately a third
(Table~\ref{table:conv_complexity}).

\begin{table*}[t]
  \centering
  \begin{tabular}{llll}
    \hline
    Pass    &Direct   &FFT-based    &FFT-based (Memoized)
    \\ \hline
    Forward &
    $f' \cdot f \cdot n'^3 \cdot k^3$ &
    $3Cn^3 \log n[f'+f+f' \cdot f] + 4f' \cdot f \cdot n^3$ &
    $3Cn^3 \log n[f'+f+f' \cdot f] + 4f' \cdot f \cdot n^3$

    \\
    Backward &
    $f' \cdot f \cdot n'^3 \cdot k^3$ &
    $3Cn^3 \log n[f'+f+f' \cdot f] + 4f' \cdot f \cdot n^3$ &
    $3Cn^3 \log n[f'+f] + 4f' \cdot f \cdot n^3$

    \\
    Update &
    $f' \cdot f \cdot n'^3 \cdot k^3$ &
    $3Cn^3 \log n[f'+f+f' \cdot f] + 4f' \cdot f \cdot n^3$ &
    $3Cn^3 \log n[f' \cdot f] + 4f' \cdot f \cdot n^3$

    \\
    Total &
    $3f' \cdot f \cdot n'^3 \cdot k^3$ &
    $9Cn^3 \log n[f'+f+f' \cdot f] + 12f' \cdot f \cdot n^3$ &
    $6Cn^3 \log n[f'+f+f' \cdot f] + 12f' \cdot f \cdot n^3$

    \\ \hline

  \end{tabular}
  \caption{Computational complexity of a fully connected convolutional
    layer, which maps $f$ input images to $f'$ output images using
    $ff'$ kernels. FFT complexity for an $n\times n\times n$ image is
    assumed to be $Cn^3\log n^3$.  Complexity is measured in number of
    floating point operations.
  \label{table:conv_complexity}
  }
\end{table*}


%

\section{Task dependency graph}

\begin{figure*}[tb]
\begin{center}
  \includegraphics[width=0.8\textwidth]{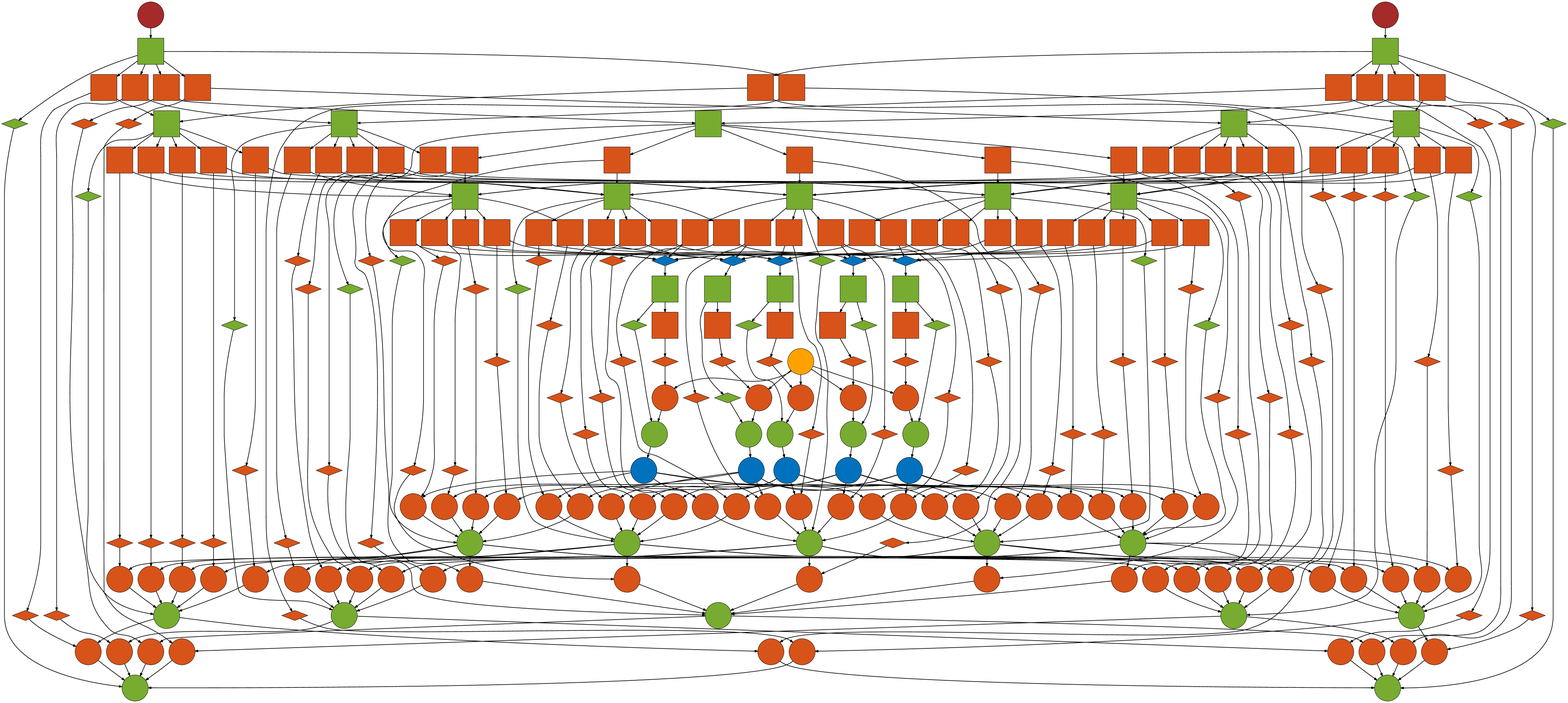}
  \end{center}
  \caption{Task dependency graph. Each edge of the computation graph
    (Fig.~\ref{fig:convnet_dag}) generates multiple nodes of the task
    dependency graph, corresponding to forward (circle), backward
    (square), and update (diamond) tasks. Node colors indicate
    transfer function (green), convolution (red), pooling/filtering
    (blue), input provider (orange), loss gradient (dark red).
   \label{fig:task_deps}
}
\end{figure*}

The entire gradient learning calculation can be represented by the
task dependency graph (Fig. \ref{fig:task_deps}).  Each node
represents one of the four forward operations (convolution,
max-pooling/filtering, transfer function), four backward operations
(Jacobians), or two update operations (kernel, bias) described above.
Two additional tasks interface with the training set.  The {\bf data
  provider} obtains a training sample used for a single round of
training, and the {\bf loss gradient} calculates the gradient of the
loss with respect to the network output.

The edges of the task dependency graph represent dependencies.  The
forward task of an edge $e = (u,v)$ in the computation graph depends
on forward pass tasks of all edges $(w,u)$.  The backward task of the
same edge depends on the backward tasks of all edges $(v,w)$.  Finally
the update task of an edge depends on both forward and backward tasks
of the same edge.

Additionally, if there was a backward pass executed before the current
forward pass, the forward task of $e$ also depends on the previous
update task of $e$.  This is relevant because gradient learning is
iterative, so the gradient is calculated repeatedly by cycling through
forward, backward, and update.

Fig.~\ref{fig:task_deps} shows the ConvNet learning graph
corresponding to the ConvNet computation graph of
Fig.~\ref{fig:convnet_dag}.  For convenience of analysis, steps
$3-5$ of one iteration of gradient learning are followed by steps $1$
and $2$ of the next iteration.

Therefore forward tasks are at the bottom of the graph, and backward
tasks at the top.  The topmost dark red circle nodes represent the
tasks that calculate the gradient of the loss with respect to the
output of the network obtained in the previous forward pass. The
yellow circle in the middle represents the task providing the input
image for the forward tasks at the bottom.  Note that there are no
update tasks for pooling/filtering.

\subsection{Theoretically achievable speedup}

Define $T_P$ as the time required for $P$ processors to perform one
learning iteration.  We would like a parallel algorithm that achieves
a large speedup $S_P=T_1/T_P$, and ideally one that approaches linear
speedup, $S_P = P$.  This should be possible for ``wide'' ConvNet
architectures, which contain many convolutions that can be done in
parallel.  We formalize this intuition in the following.

According to Brent's theorem~\cite{bretttheorem}, if a computation can
be performed in $T_\infty$ time with an infinite number of processors,
then
\begin{equation}\label{brents_theorem}
  T_P \le T_{\infty} + \frac{T_1 - T_{\infty}}{P}
\end{equation}
This amounts to a speedup of at least
\begin{equation}\label{brents_theorem_2}
  S_P \equiv \frac{T_1}{T_P} \ge \frac{S_\infty}{1 + \frac{S_\infty - 1}{P}}
\end{equation}
We will refer to the right hand side as the ``theoretically achievable
speedup,'' because it depends on the idealized assumptions of the PRAM
model used to prove Brent's theorem.

We will estimate the theoretically achievable speedup for layered
architectures in which every convolutional layer is fully connected.
As before, time complexity is measured in number of floating point
instructions.  We can already estimate $T_1$ by summing the times in
Tables \ref{table:other_complexity} and \ref{table:conv_complexity}
for each layer of the network.  To estimate $T_\infty$, we analyze the
following algorithm employing an infinite number of processors.  (1)
Move sequentially through the layers, and perform all forward tasks in
each layer in parallel.  (2) Compute the loss gradient for all output
nodes in parallel.  (3) Move sequentially backward through the layers,
and perform all backward tasks in each layer in parallel.  (4) Perform
the weight updates for all kernels and biases in parallel.

Since the layers are done sequentially, the total time for the forward
pass is the sum of contributions from each layer (convolutional,
transfer function, or max pooling/filtering) as specified in
Tables~\ref{table:t_inf_conv}~and~\ref{table:t_inf_other}.  The time
for the backward pass is calculated similarly.  Since all kernel and
bias updates are done in parallel, the total update time is the
maximum of the individual update times, as specified in
Tables~\ref{table:t_inf_conv}~and~\ref{table:t_inf_other}.  The sum of
forward, backward, and update times yields the time complexity of one
gradient learning iteration.

Most of the formulas in the tables do not depend on the widths of the
layers, $f$ and $f'$.  This is because all tasks in a layer are done
in parallel.  The only exception is that the complexity of a
convolutional layer depends logarithmically on width ($\ceil{ \log_2
  f}$), because summing the results of $f$ convergent convolutions
requires this amount of time using the binary collapse algorithm
described in \cite{bretttheorem}.

\begin{figure}
  \centering
  \subfloat[]{\protect\includegraphics[width=0.24\textwidth]
    {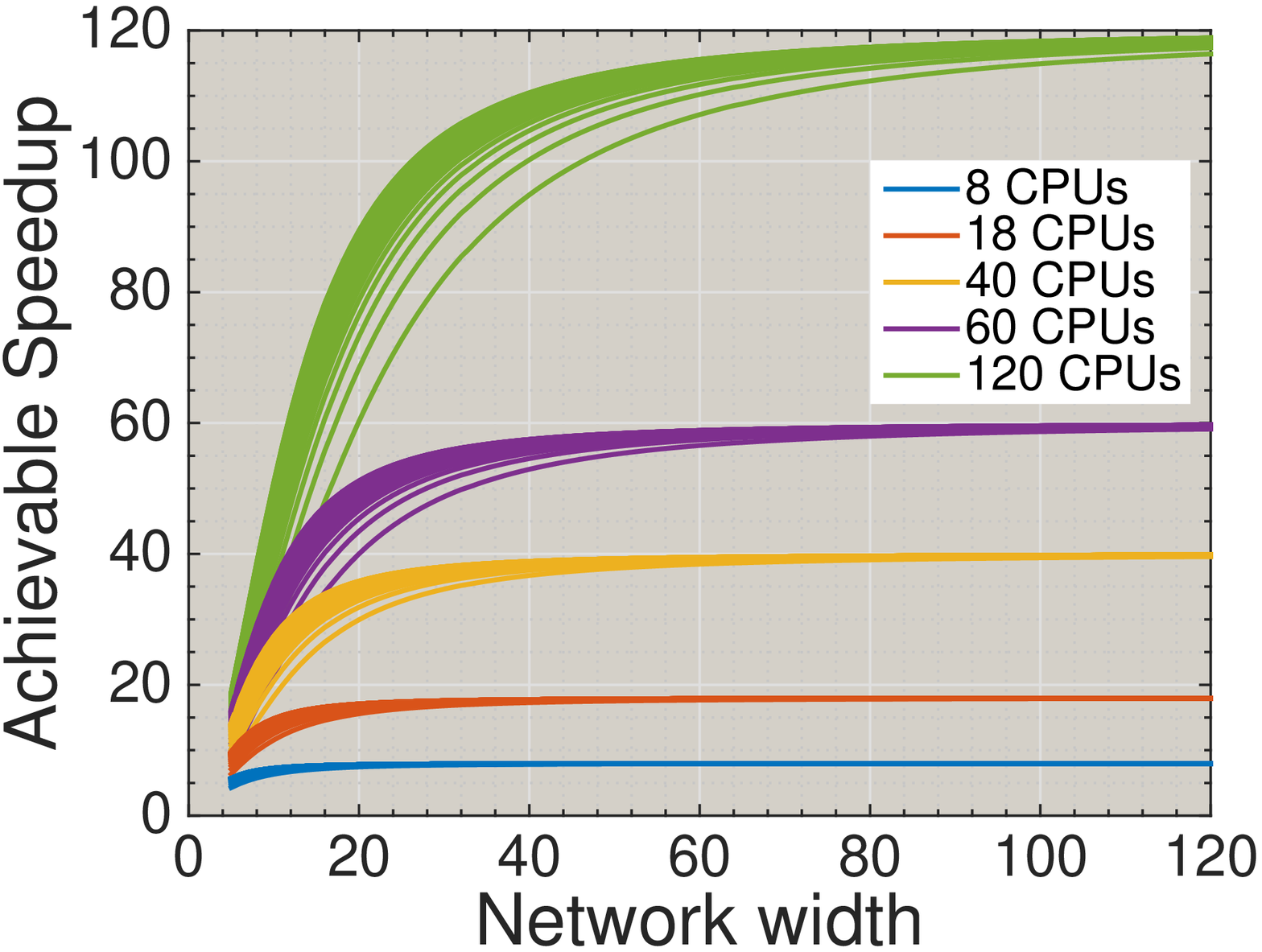}}
  \subfloat[]{\protect\includegraphics[width=0.24\textwidth]
    {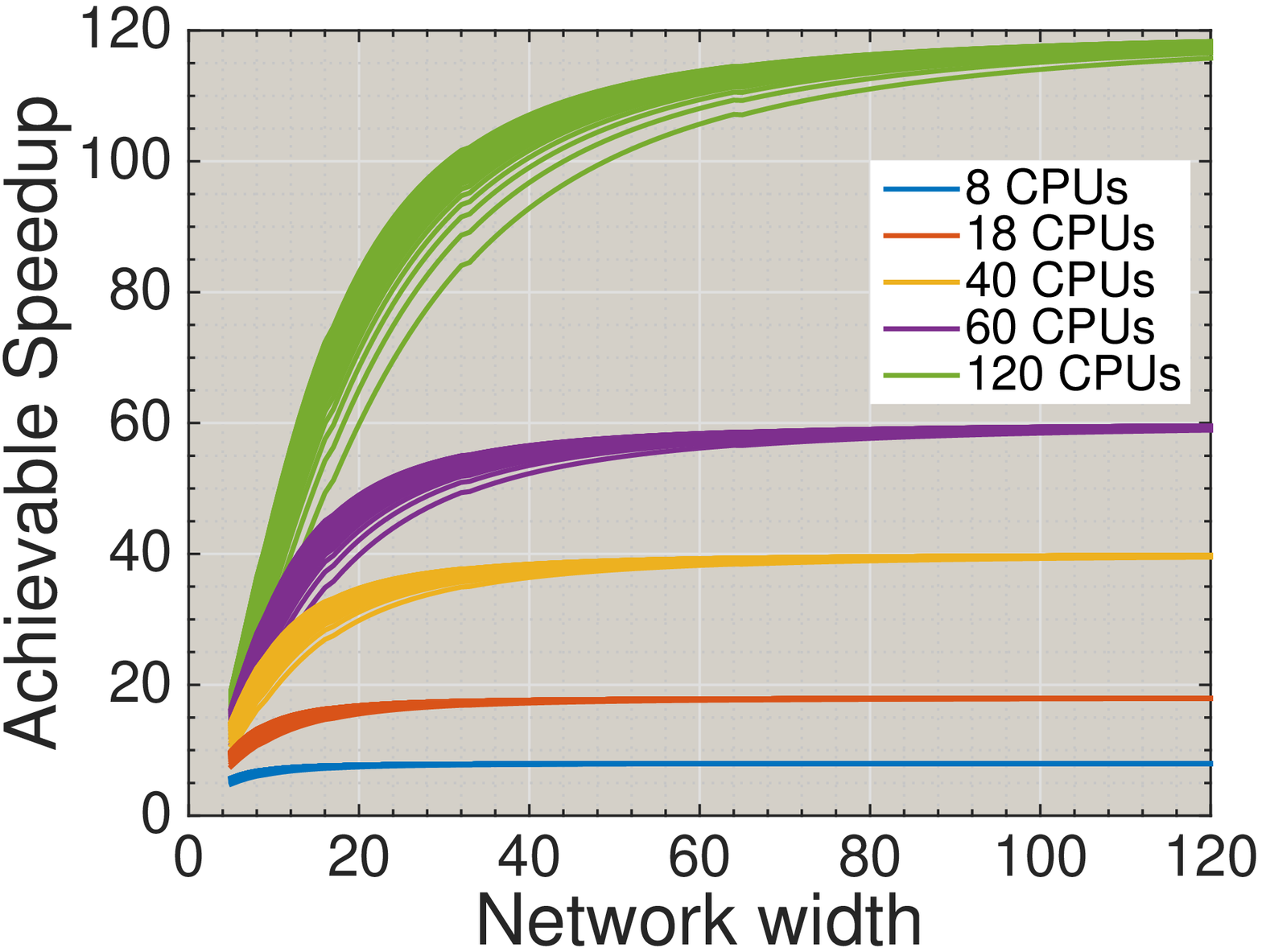}}
  \caption{Theoretically achievable speedup (\ref{brents_theorem_2})
    using (a) direct convolution (b) FFT-based convolution with
    memoizing enabled.  Multiple lines of the same color represent
    networks with different depths, ranging from 4 to 40.  The kernels
    in all the networks have size of $5^3$.
  \label{fig:achievable_speedup}}
\end{figure}

Plots of the theoretically achievable speedup (\ref{brents_theorem_2})
for networks of different width and depth are shown in
Fig.~\ref{fig:achievable_speedup}\footnote{The constant $C$ for the FFT
operations is assumed to be 5.}.  In all cases, $S_P\to P$ in the limit
of large network width $f$.  This is because $T_1$ scales like $f^2$
for large $f$ (see terms in Table \ref{table:conv_complexity}), while
$T_\infty$ scales like $\log f$ (see terms in Table
\ref{table:t_inf_conv}).  It follows that $S_\infty$ diverges with
$f$, so the bound on $S_P$ in Eq. (\ref{brents_theorem_2}) is equal to
$P$ in the limit of large $f$.

According to Fig.~\ref{fig:achievable_speedup}, the network width at
which $S_P$ reaches a fixed fraction (say 75\%) of its maximal value
($P$), increases with $P$. This behavior is consistent with
Eq. (\ref{brents_theorem_2}).  We expect $S_P$ to approach a fixed
fraction of $S_\infty$ when $S_\infty \approx P$.  Since $S_\infty$
scales like $f^2$ (neglecting the logarithmic factor due to
$T_\infty$), this should happen when $f^2\approx P$.  The power of two
means that the theoretically achievable speedup approaches its maximum
value even for networks with rather modest widths.

\begin{table*}[t]
  \centering
  \begin{tabular}{llll}
    \hline
    Pass    &Direct   &FFT-based    &FFT-based (Memoized)
    \\ \hline
    Forward -- $T_{\infty}^{fwd}$ &
    $n'^3 \cdot k^3 + n'^3 \ceil*{\log_2 f}$ &
    $6Cn^3 \log n + 4n^3 \ceil*{\log_2 f}$ &
    $6Cn^3 \log n + 4n^3 \ceil*{\log_2 f}$
    \\
    Backward -- $T_{\infty}^{bwd}$ &
    $n'^3 \cdot k^3 + n^3 \ceil*{\log_2 f'}$ &
    $6Cn^3 \log n + 4n^3 \ceil*{\log_2 f'}$ &
    $6Cn^3 \log n + 4n^3 \ceil*{\log_2 f'}$
    \\
    Update -- $T_{\infty}^{update}$&
    $n'^3 \cdot k^3$ &
    $6Cn^3 \log n + 4n^3$ &
    $3Cn^3 \log n + 4n^3$
    \\ \hline

  \end{tabular}
  \caption{Time required to perform operations on fully connected
    convolutional layers with infinite number of processors
    available.}
  \label{table:t_inf_conv}
\end{table*}

\begin{table}
  \centering
  \begin{tabular}{llll}
    \hline
    Pass    &Pooling   &Filtering    &Transfer function
    \\ \hline
    Forward -- $T_{\infty}^{fwd}$ & $n^3$ & $6n^3 \log k$ & $n^3$
    \\
    Backward -- $T_{\infty}^{bwd}$ & $n^3$ & $n^3$ & $n^3$
    \\
    Update -- $T_{\infty}^{update}$ & $-$ & $-$ & $n^3$
    \\ \hline
  \end{tabular}
  \caption{Time required to perform pooling, filtering and transfer
    function on a full layer with infinite number of processors
    available.}
  \label{table:t_inf_other}
\end{table}


\section{Task scheduling and execution}

Brent's theorem guarantees the existence of a parallel algorithm that
achieves a large speedup for training wide ConvNets.  We now turn to
the problem of designing a parallel algorithm that actually achieves
large speedup in practice.  Since Brent's theorem assumes no
synchronization and communication overhead, we design our algorithm to
minimize synchronization overhead and increase temporal locality of
computation to reduce cache misses.

The central quantity in our algorithm is a global priority queue that
contains tasks that are ready to be executed together with their
priority.  A predetermined number of workers will then execute the
tasks from the global queue.

\subsection{Priority queue}

Tasks are placed on a global queue when all non-update dependencies
are satisfied (only tasks with update task as requirements are forward
tasks).  The rationale behind this design choice is that if a forward
task is scheduled for execution without the required update task being
done, we will force execution of the update task followed by the
forward task that requires the result of the update, hence increase
the memory locality.

The tasks on the queue are sorted by priority.  Priorities are chosen
to increase temporal locality of the computation and minimize the
latency of the computation.  We introduce two unique strict orderings
of the nodes in the ConvNet's computation graph based on the longest
distance, in decreasing order, to any output and input node
respectively.  Nodes with the same distance will be ordered in some
unique way.  The priority of the forward task of an edge $e = (u,v)$
will be equal to position of the output node $v$ in the ordering based
on the distance to the output nodes, and similarly the priority of the
backward task will be equal to the ordering of $u$ based on the
distance to the input nodes.  This ensures that we prioritize tasks
with the longest path to a sink node in the task dependency, which
should intuitively favor lower latency schedules.  The strict ordering
of the tasks with the same distance increases temporal locality by
assuring that when multiple tasks with the same distance are scheduled
we prefer to execute ones computing 3D images that have to be
accumulated in the same sum, thus increasing the probability of the
memory accessed being in the cache.

The update tasks will have the lowest priority of all tasks.  Their
execution will be forced when their result is required for the forward
pass, which increases cache locality as the result will be used
immediately.  The only other time the update tasks will be executed is
if there's no other forward or backward tasks ready to be executed.

\subsection{Task Execution}

The tasks are executed by $N$ workers.  Each worker picks up and
executes a task with the highest priority on the queue.


{\bf Forward task} algorithm is shown in the
Algorithm~\ref{alg:forward_task}.  The main functionality of the
forward task is to apply the appropriate $\proc{Forward-Transform}$ on
the given input image $I$ and accumulate the result to the sum stored
in the output node.  The task that adds the last image to the sum then
queues all dependent forward tasks for execution.

The main functionality of the forward task is shown in the procedure
$\proc{Do-Forward}$.  However such procedure can only be executed when
the update task from the previous round has been completed.  This is
ensured by creating a new sub-task containing the main functionality
and calling the $\proc{Force}$ method.

\begin{algorithm}
  {\small
  \begin{codebox}
    \Procname{$\proc{Forward-Task}(e,I)$}
    \li $t \gets \proc{Create-Task}(\proc{Do-Forward},e,I)$
    \li $\proc{Force}(\attrib{e}{update\_task}, t)$
  \end{codebox}

  \begin{codebox}
    \Procname{$\proc{Do-Forward}(e = (u,v),I)$}
    \li $I^{out} \gets \attrib{e}{\proc{Forward-Transform}(I)}$
    \li \If $\proc{Add-To-Sum}(\attrib{v}{fwd\_sum},I^{out})$
    \li     \Then $\attrib{v}{I_f} \gets \proc{Get-Sum}(\attrib{v}{fwd\_sum})$
    \li           \For $e' \in \attrib{v}{out\_edges}$
    \li            \Do $t \gets \proc{Create-Task}(\proc{Forward-Task},e',\attrib{v}{I_f})$
    \li                $\proc{Enqueue}(e'.\id{fwd\_priority}, t)$
    \End
    \End
  \end{codebox}
  }

  \caption{Executing a forward task}
  \label{alg:forward_task}
\end{algorithm}

The $\proc{Force}$ function receives an update task and a forward
subtask as parameters.  The goal of the function is to execute the
forward subtask but also make sure that the update task has been
completed.  In order to do that the method first examines the state of
the update task which can be one of the following (Note that the
$\proc{Force}$ is called from the thread scheduled to execute the
appropriate forward task).

\begin{enumerate}
\item {\bf Completed} - the execution of the update task has been
  completed; in this case the calling thread just executes the forward
  subtask.
\item {\bf Queued} - the update task is on the queue waiting to be
  scheduled for execution; in this case the update task is removed
  from the queue, and the calling thread executes both the update task
  and the forward subtask.
\item {\bf Executing} - the update task is currently being executed by
  some other thread; in this case the forward subtask gets
  \emph{attached} to the update task.  This flags the thread executing
  the update task to execute the forward subtask as soon as the update
  task is completed.  The calling thread then returns and picks up
  another task for execution.
\end{enumerate}

Such design ensures that no thread is ever waiting for completion of
an update task, but rather executes the required update task itself,
or delegates the forward subtask to the thread currently executing the
update task.

\begin{algorithm}
  {\small
  \begin{codebox}
    \Procname{$\proc{Backward-Task}(e = (u,v),I)$}
    \li $I^{out} \gets \attrib{e}{\proc{Backward-Transform}(I)}$
    \li \If $\attrib{e}{is\_trainable}$
    \li \Then $I_f = \attrib{u}{fwd\_image}$
    \li       $\attrib{e}{update\_task} \gets \proc{Create-Task}(\proc{Update},e,I_f,I)$
    \li       $\proc{Enqueue}(\id{lowest\_priority}, \attrib{e}{update\_task})$
        \End
    \li \If $\proc{Add-To-Sum}(\attrib{u}{bwd\_sum},I^{out})$
    \li     \Then $\attrib{u}{I_b} \gets \proc{Get-Sum}(\attrib{u}{bwd\_sum})$
    \li           \For $e' \in \attrib{u}{in\_edges}$
    \li            \Do $t \gets \proc{Create-Task}(\proc{Backward-Task},e',\attrib{u}{I_b})$
    \li                $\proc{Enqueue}(\attrib{e'}{bwd\_priority}, t)$
    \End
    \End
  \end{codebox}
  }
  \caption{Executing a backward task}
  \label{alg:backward_task}
\end{algorithm}

{\bf Backward task} algorithm is shown in
Algorithm~\ref{alg:backward_task}.  When scheduled for execution, all
the dependencies of the backward task have been satisfied.  The
backward task then applies the appropriate $\proc{Backward-Transform}$
on the given image and then queues an appropriate update task for
execution with the lowest common value as priority.  Similarly to the
forward task, the transformed image is then added to the sum stored in
the input node, and the thread to add the last image to the sum queues
the dependent tasks for execution.

{\bf Update tasks} algorithm is shown in
Algorithm~\ref{alg:update_task}.  First the gradient of the loss is
calculated, and then is multiplied by a small ``learning rate'' $\eta$
and subtracted from the set of the training parameters (weights of the
kernel or the bias).  Finally, if a forward subtask has been attached
it is detached and executed.

\begin{algorithm}
  {\small
  \begin{codebox}
    \Procname{$\proc{Update}(e,I_f,I_b)$}
    \li $G \gets \attrib{e}{\proc{Compute-Gradient}}(I_f,I_b)$
    \li $\attrib{e}{params} \gets \attrib{e}{params} - \attrib{e}{\eta} \cdot G$
    \li \If $\attrib{this}{fwd\_subtask}$
    \li     \Then $t \gets \attrib{this}{fwd\_subtask}$
    \li           $\attrib{this}{fwd\_subtask} \gets \id{NIL}$
    \li           $\proc{Execute}(t)$
    \End
  \end{codebox}
  }
  \caption{Executing an update task}
  \label{alg:update_task}
\end{algorithm}

\section{Synchronization issues}

It is important to minimize the amount of time spent in critical
sections -- parts of the code that can be only executed by a single
thread at a time.
The main three points in the algorithm that require synchronization
are memory management (allocation/deallocation), operations on the
global task queue and concurrent summations.

\subsection{Queue operations}

The operations on the global task priority queue have to be
synchronized.  The queue is implemented as a heap of lists lowering
the complexity of insertion and deletion from $\log N$ to $\log K$,
where $N$ is the total number of tasks in the queue and $K$ is the
number of distinct values for the priority of the tasks inside the
queue.  Depending on the network structure, this number can be much
smaller than the total number of tasks in the queue, which is
especially true for wide networks.

\subsection{Wait-free concurrent summation}

When multiple edges converge on the same node in the computation
graph, it means that multiple convolutions executed in parallel need
to add their results to the same accumulated sum.  The additions have
to be synchronized; only one thread is allowed to change the sum.  The
naive strategy, waiting until all other threads have finished adding
their images to the sum, would lead to critical section time that
scales linearly with the image size $n^3$.  We propose a novel method
that eliminates the dependence on image size by performing only
pointer operations inside the critical section, which works as
follows.

Suppose that multiple threads are executing $\proc{Add-To-Sum}$ in
Algorithm~\ref{alg:sum}.  For each thread, $v$ points to a different
3D image. We would like the pointer to the sum of all these images to
be stored in the object $S$ when the computation terminates.  This is
accomplished by having each thread repeatedly try to reset the pointer
to the sum stored in $\attrib{S}{sum}$ to point to $v$ instead.  If
the thread succeeds, it stops working.  If the thread fails, it adds
the value pointed to by $\attrib{S}{sum}$ to the location referenced
by $v$, and sets the pointer to $\const{NIL}$.  Every thread continues
to work until it succeeds.  Once the last thread succeeds, $S$ will
contain the correct answer.  Note that this algorithm does the
time-consuming additions outside the critical section (lines 5-11).

\begin{algorithm}
  {\small
  \begin{codebox}
    \Procname{$\proc{Add-To-Sum}(S,v)$}
    \li $v' \gets \const{nil}$
    \li $last \gets \const{false}$
    \li \While $\const{TRUE}$ \Do
    \li \proc{Acquire}($\attrib{S}{lock}$) \Do
    \li     \If   $\attrib{S}{sum} \isequal \const{nil}$
    \li     \Then $\attrib{S}{sum} \gets v$
    \li           $v \gets \const{nil}$
    \li           $\attrib{S}{total} \gets \attrib{S}{total} + 1$
    \li           $last \gets (\attrib{S}{total} \isequal \attrib{S}{required})$
    \li   \Else $v' \gets \attrib{S}{sum}$
    \li         $\attrib{S}{sum} \gets \const{nil}$
          \End 
        \End 
    \li \proc{Release}($\attrib{S}{lock}$)
    \li \If   $v \isequal \const{nil}$
    \li \Then \Return $last$
    \li \Else $\proc{Add-To}(v,v')$ \Comment $v \gets v + v'$
        \End 
        \End 
   \end{codebox}


  }
  \caption{Wait-free concurrent summation algorithm}
  \label{alg:sum}
\end{algorithm}

\subsection{Memory management}

ZNN implements two custom memory allocators.  These are designed to be
faster than standard memory management routines, at the cost of using
more memory. One custom allocator is dedicated to 3D images, which are
usually large, and the other is dedicated to small objects used in
auxiliary data structures. Both allocators maintain $32$ global pools
of memory chunks. Each pool $i$, $i \in 0\dots 31$ contains chunks of
sizes of $2^i$. Lock-free queues, as described in
~\cite{michael1996simple} and implemented as a part of the
boost~\cite{boost_lockfree} library are used to implement the pool
operations. The only difference between the allocators is the memory
alignment---the 3D image memory allocator ensures proper memory
alignment for utilizing SIMD instructions.  No memory is shared
between the two allocators.

When a chunk of memory of size $s$ is requested, first $s$ is rounded
up to the nearest power of $2$.  The appropriate pool is examined for
available memory chunks.  If there's an available chunk we return it
and remove it from the pool. If no chunks are available we allocate
one from the system and return it.

When de-allocating a chunk memory, it is simply added to the
appropriate pool, and no memory is ever returned to the system. This
means that the memory usage of our program can never decrease.  In
practice, as the ConvNet training consist of a single loop performing
the same work, our memory usage peaks after a few rounds.

In the worst case this strategy can lead to near $2\times$ memory
usage overhead; however the available memory to the CPU is rarely a
limiting factor in training a network.  In the future, we might
consider implementing more advanced memory allocators, such as ones
with thread-local pools in addition to the global pool, or ones with
higher granularity of available chunk sizes to reduce the size
overhead.



\section{Scalability}

We performed measurements of the speedup achieved by our proposed
parallel algorithm relative to the serial algorithm, using the CPU
systems listed in Table~\ref{table:test_machines}.  Amazon EC2
instances with $8$ and $18$ cores (c4.4xlarge and c4.8xlarge) were
chosen for benchmarking, because they are readily available to anyone.
A 4-way CPU system was included because it has $40$ cores, though this
is a relatively specialized piece of hardware.  For an even larger
number of cores, we also benchmarked the Xeon Phi\texttrademark
Knights Corner.  All measurements used the Intel compiler (version
15.0.2) with Intel MKL (version 11.2) libraries for FFTs and direct
convolution.

\begin{table}
  \centering
  \begin{tabular}{lll}
    \hline
    CPU & Frequency & Cores/Threads \\
    \hline
    Intel\textregistered ~Xeon\texttrademark ~E5-2666 v3 & $2.9$ GHz &
    $8$ cores/$16$ threads \\
    \hline
    Intel\textregistered ~Xeon\texttrademark ~E5-2666 v3 & $2.9$ GHz &
    $18$ cores/$36$ threads \\
    \hline
    Intel\textregistered ~Xeon\texttrademark ~E7-4850 & $2.0$ GHz &
    $40$ cores/$80$ threads \\
    \hline
    Intel\textregistered Xeon Phi\texttrademark 5110P & $1.053$ GHz &
    $60$ cores/$240$ threads \\
    \hline

  \end{tabular}

  \caption{Machines used for the experiments}
  \label{table:test_machines}
\end{table}

The 3D ConvNets contained four fully-connected convolutional (C)
layers with $3\times 3\times 3$ kernels, each followed by a transfer
function layer (T) with rectified linear function, and two $2\times
2\times 2$ max-filtering (M) layers.  Each convolutional layer The
sequence of layer types was CTMCTMCTCT. The output patch size was
$12\times 12\times 12$.

The 2D ConvNets contained 6 fully-connected convolutional layers with
$11\times 11$ kernels, each followed by rectified linear transfer
function layer (T), and two $2\times 2$ max-filtering layers (2nd and
4th).  The sequence of layer types was CTMCTMCTCTCTCT.  The output
patch size was $48 \times 48$.


The ZNN measurements were performed by first running the gradient
learning algorithm for 5 warm-up rounds and then averaging the time
required for the next 50 rounds.  The GPU measurements were averaged
over 100 rounds.

2D ConvNets were implemented as a special case of 3D ConvNets, by
setting one of the dimensions to have size one.  The width of the
ConvNets was varied as described below.  FFT convolution was employed
for 2D, and direct convolution for 3D to illustrate the use of both
methods; reversing this yields similar results.  Other network
architectures and kernel sizes also yield similar results.


\begin{figure*}
  \centering
  \subfloat[]{\protect\includegraphics[width=0.24\textwidth]
    {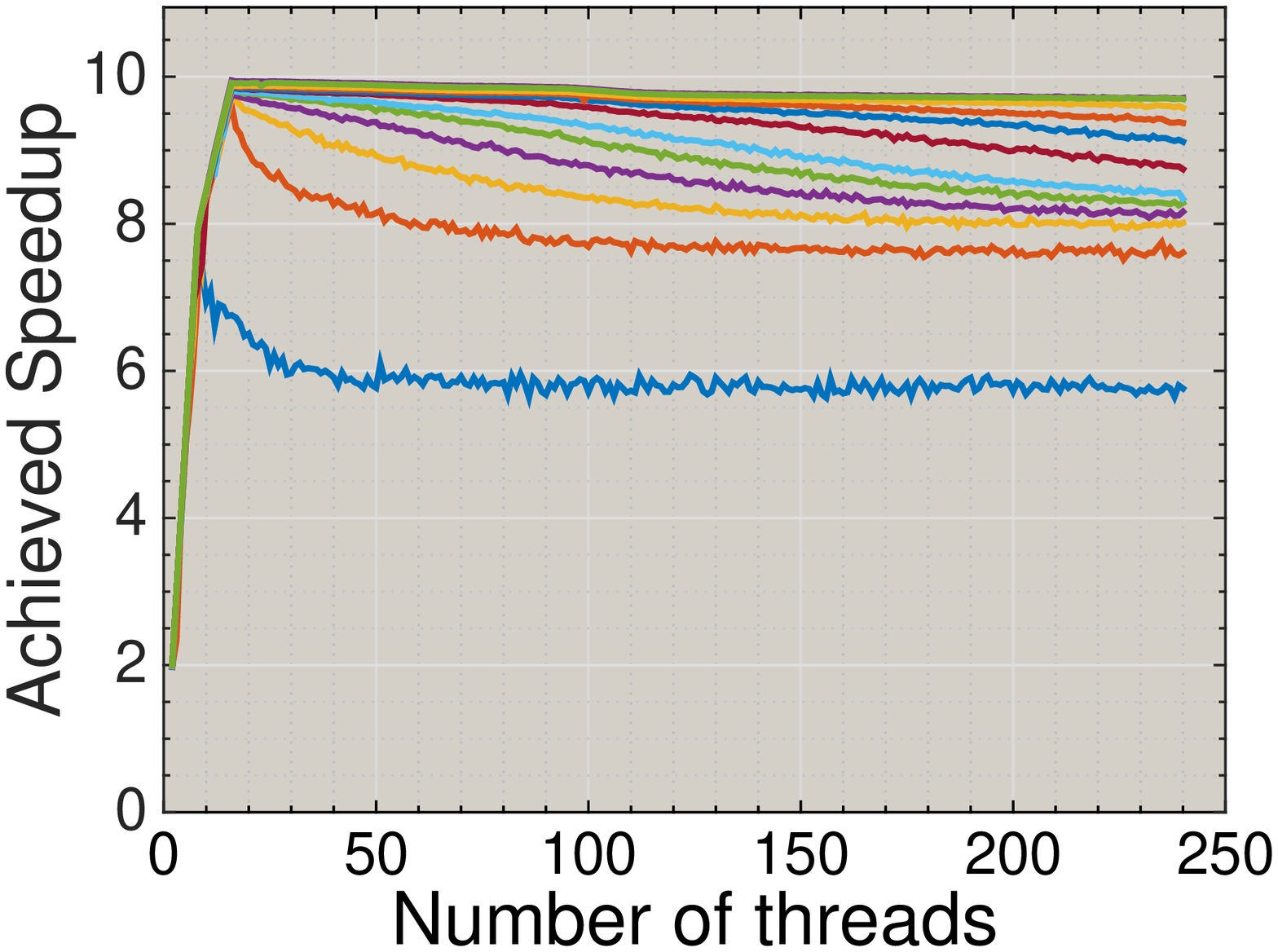}}
  \subfloat[]{\protect\includegraphics[width=0.24\textwidth]
    {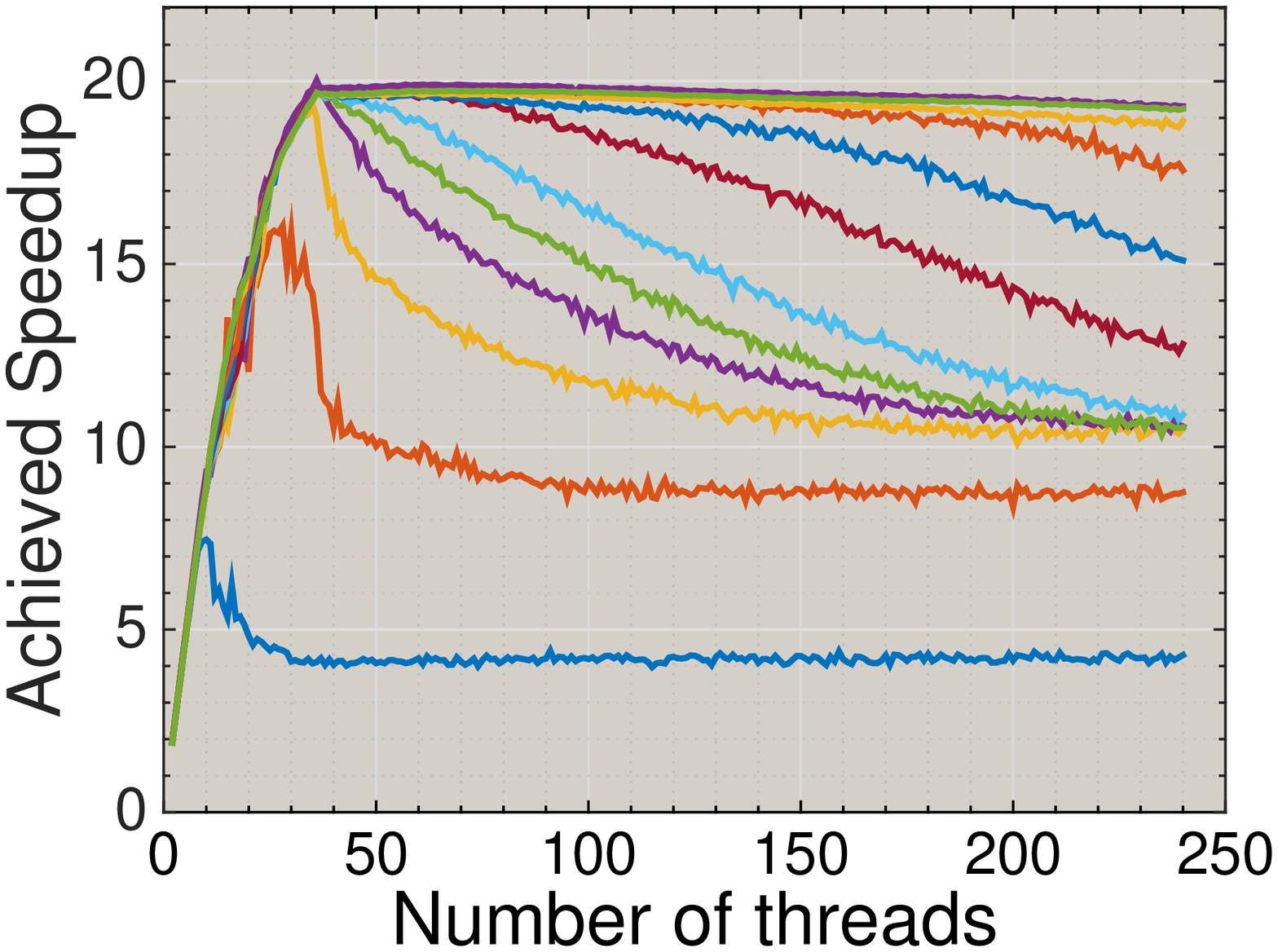}}
  \subfloat[]{\protect\includegraphics[width=0.24\textwidth]
    {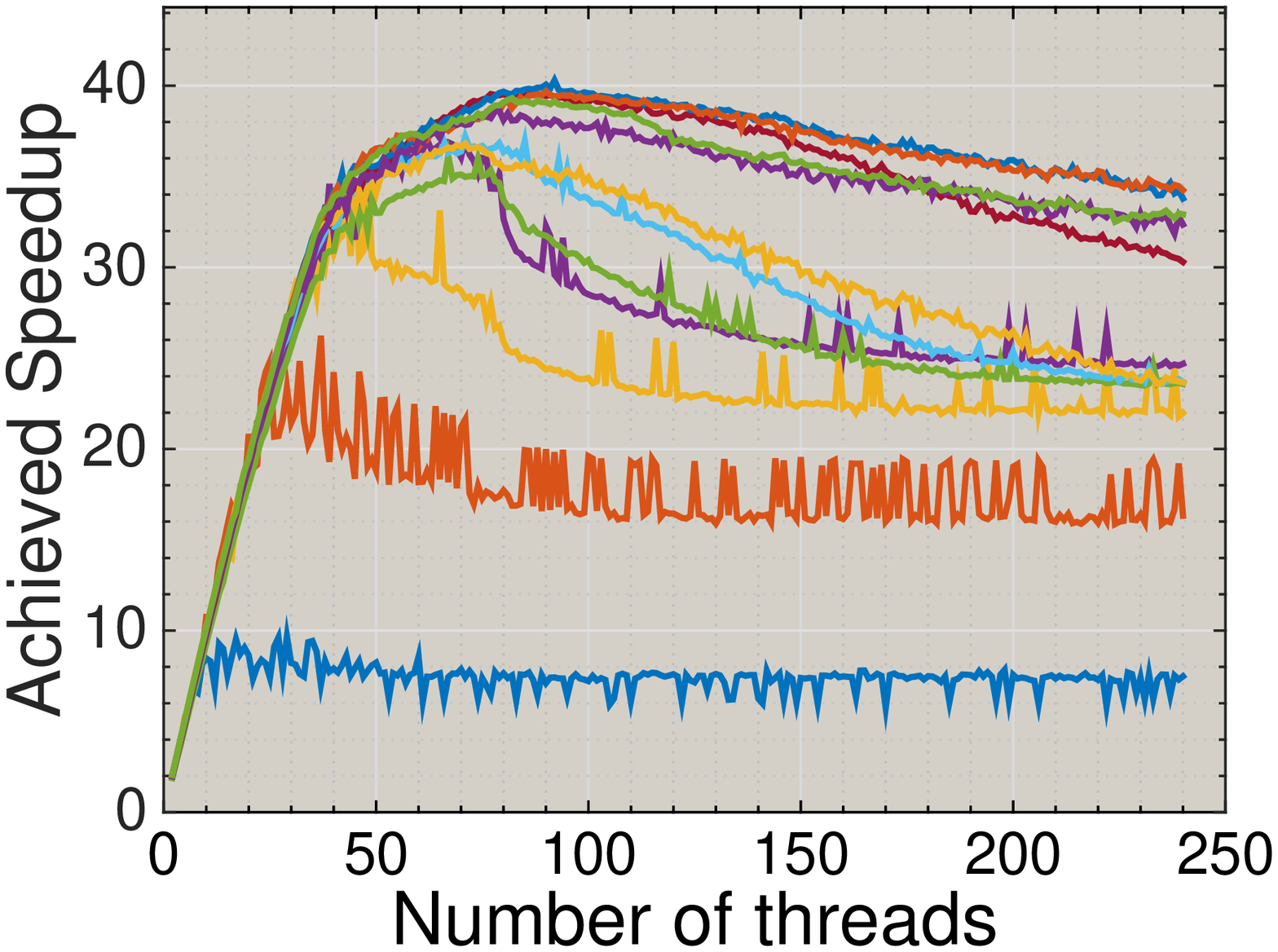}}
  \subfloat[]{\protect\includegraphics[width=0.24\textwidth]
    {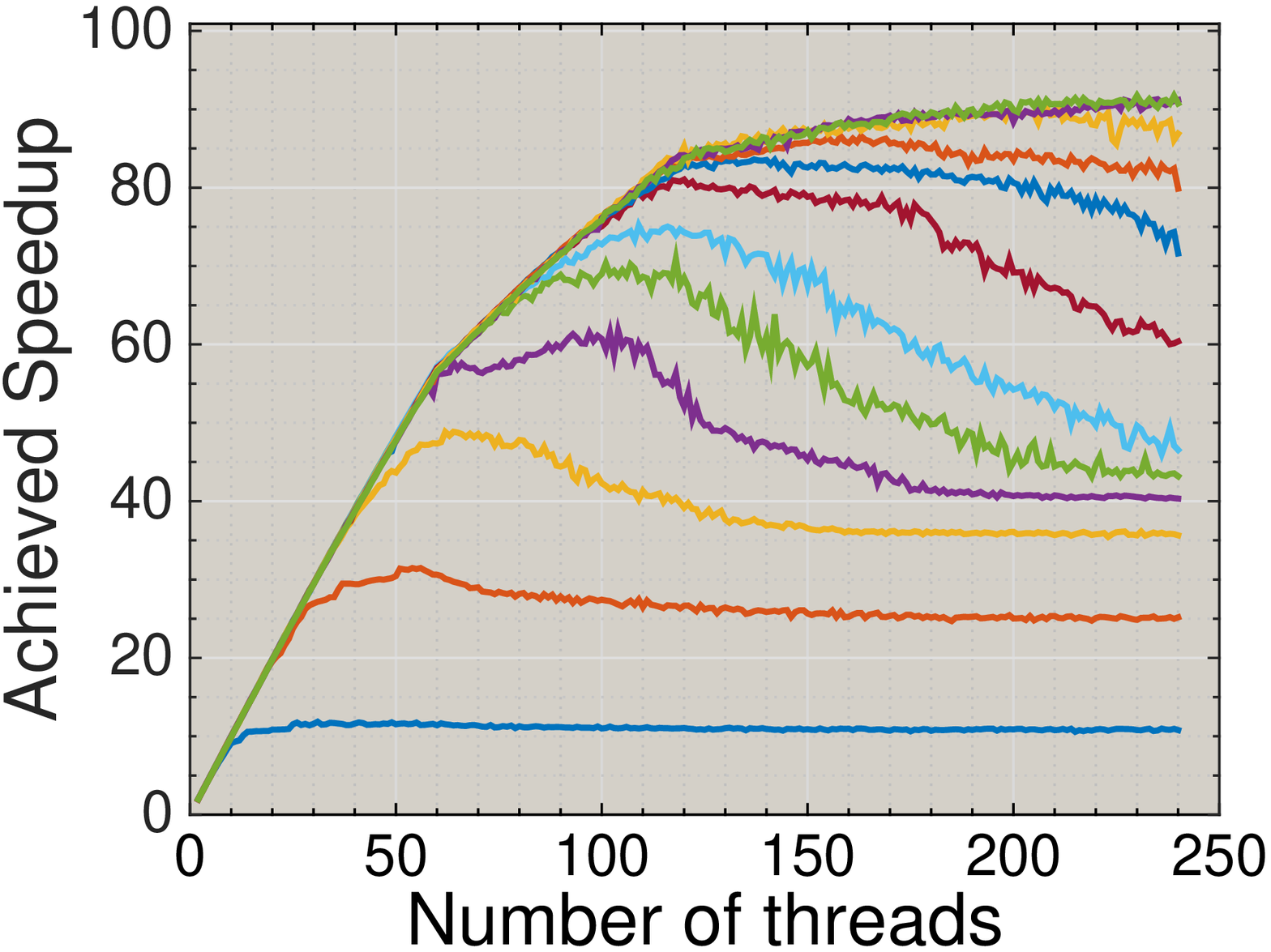}}
    \\
  \subfloat[]{\protect\includegraphics[width=0.24\textwidth]
    {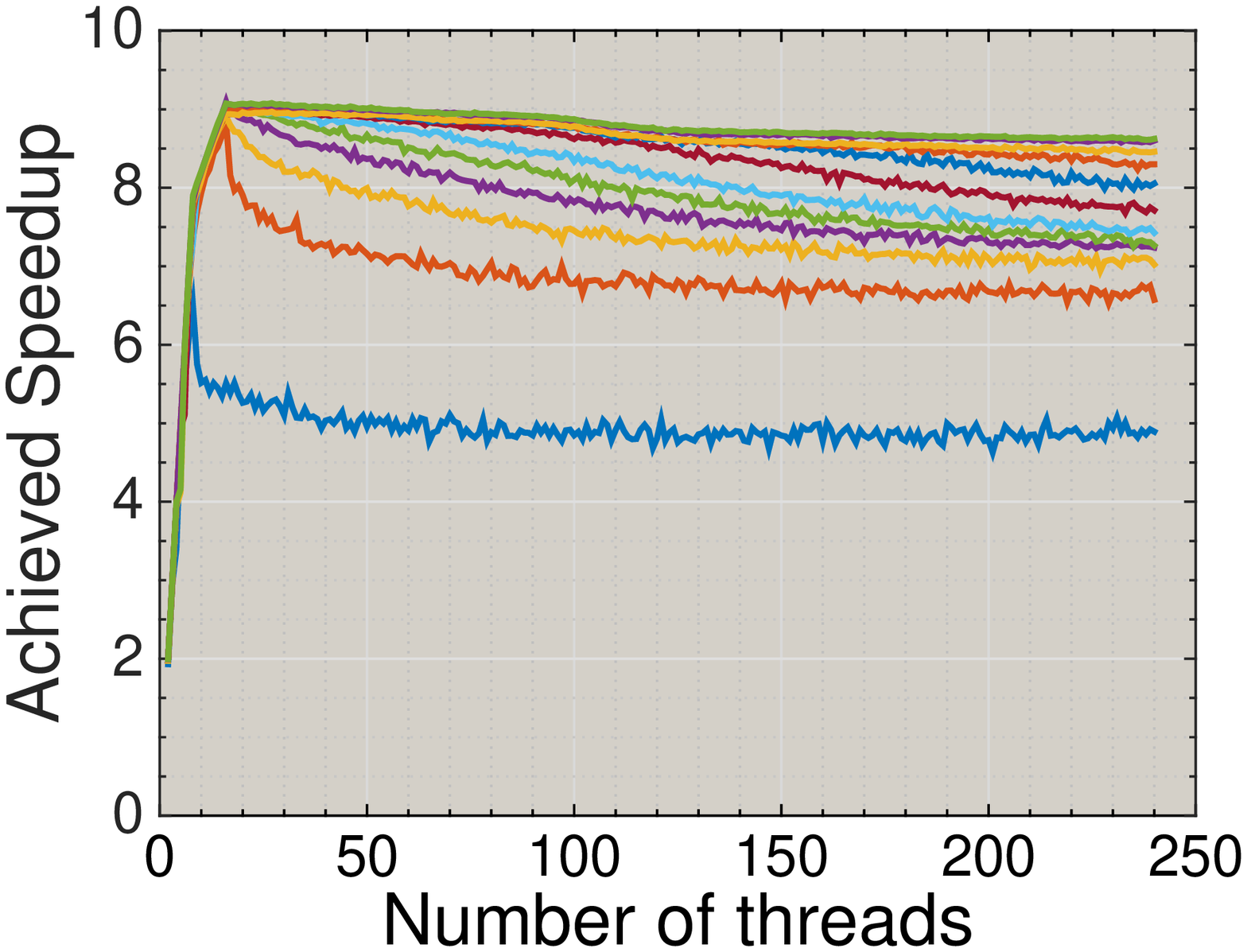}}
  \subfloat[]{\protect\includegraphics[width=0.24\textwidth]
    {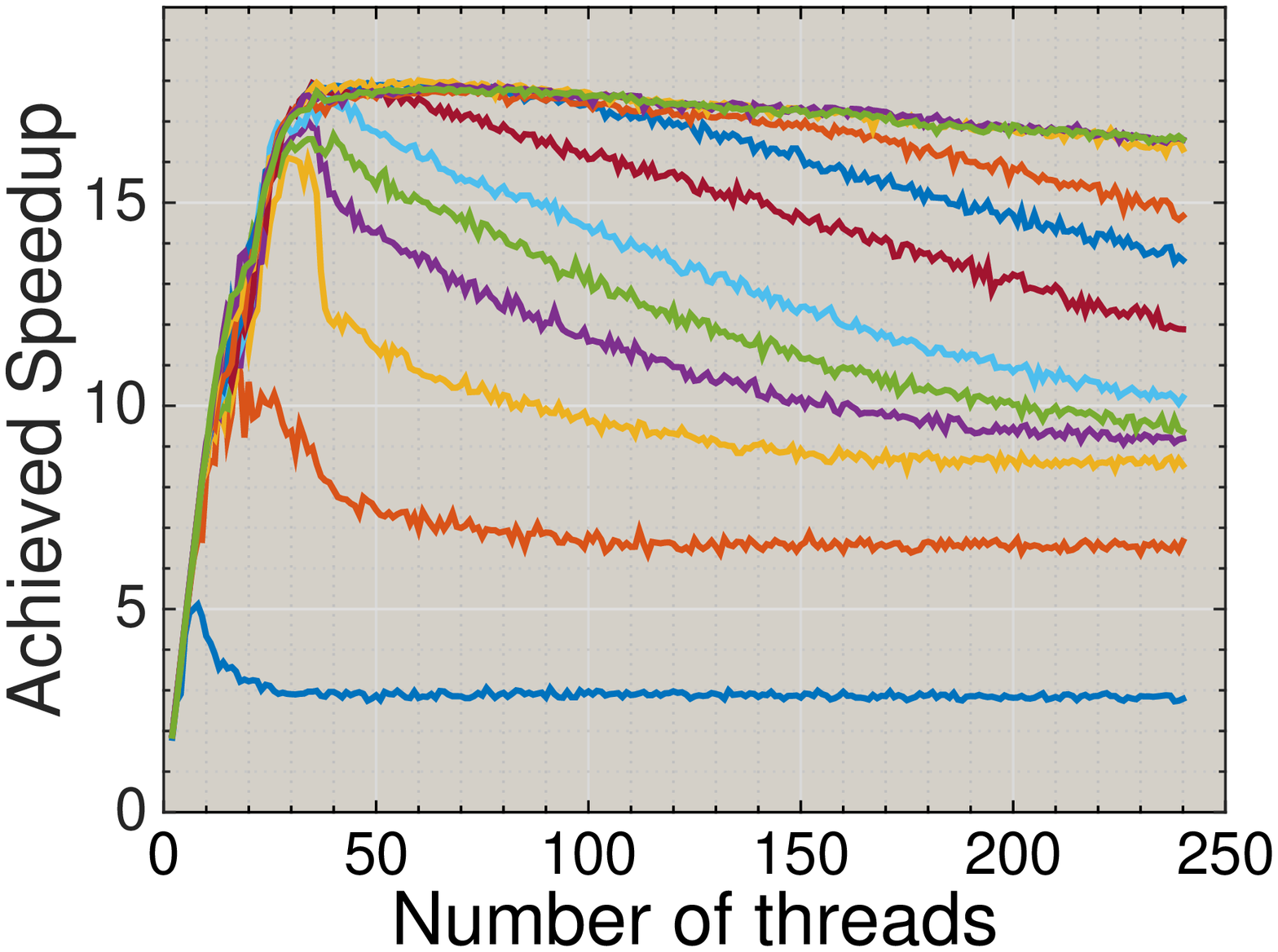}}
  \subfloat[]{\protect\includegraphics[width=0.24\textwidth]
    {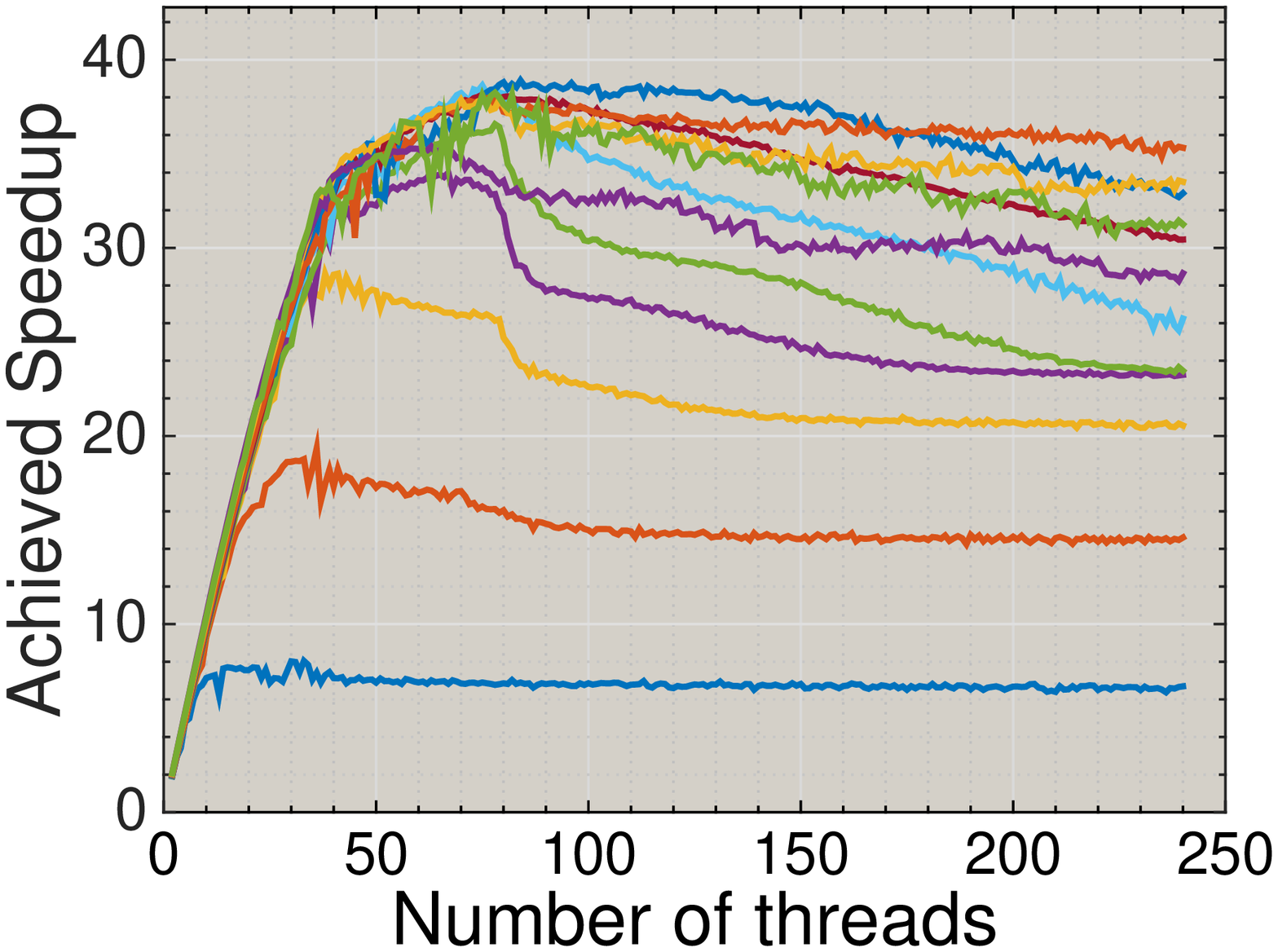}}
  \subfloat[]{\protect\includegraphics[width=0.24\textwidth]
    {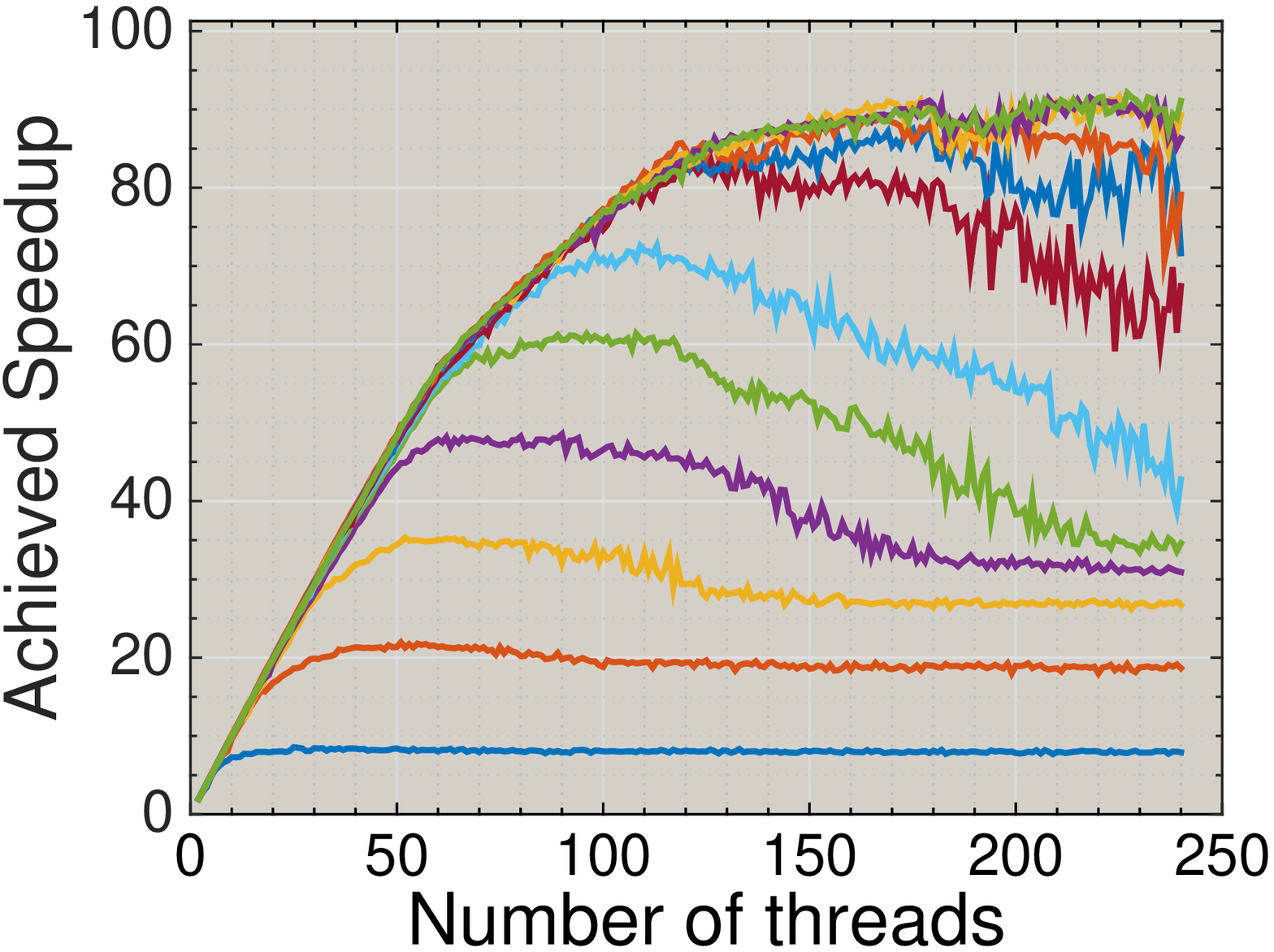}}

  \caption{Speedup versus number of threads for 2D (first row) and 3D
    (second row) achieved on machines in Table
    \ref{table:test_machines} (left-to-right columns).  In each graph,
    the lines are for network widths
    $5,10,15,20,25,30,40,50,60,80,100,120$, from bottom to top.
  }
  \label{fig:2dspeedups_threads}
\end{figure*}

Fig.~\ref{fig:2dspeedups_threads} shows speedup attained by various
CPUs as a function of two parameters, number of worker threads and
network width.  Each graph shows the result of varying the number of
workers while network width is held fixed.  To achieve near maximal
possible speedup ZNN requires sufficiently wide networks ($\geq 30$
for multicore CPUs and $\geq 80$ for the manycore CPU) and
sufficiently many worker threads (number of hyperthreads for multicore
and number of hardware threads for manycore)~\footnote{Xeon
  Phi\texttrademark has hardware threads which differ from virtual
  thread technology of the desktop Xeon\texttrademark processors.}.
The value of the maximal speedup is equal to the number of cores or a
bit larger (maximal height of graphs).

\begin{figure}
  \centering
  \includegraphics[width=0.41\textwidth]{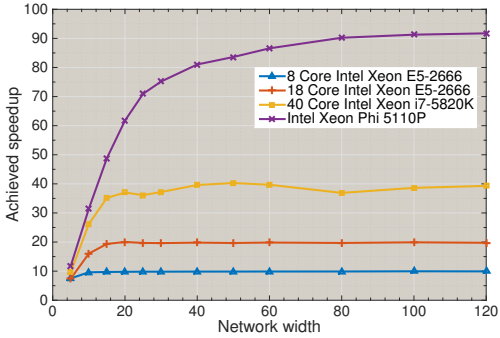}
  \caption{ Achieved speedups on 2D networks compared to the serial
    algorithm.}  \label{fig:2dspeedups}
\end{figure}

\begin{figure}
  \centering
  \includegraphics[width=0.41\textwidth]{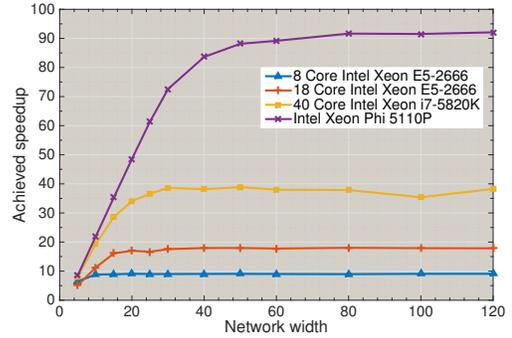}
  \caption{Achieved speedups on 3D networks compared to the serial
    algorithm.}
  \label{fig:3dspeedups}
\end{figure}

For a wide network on multicore CPUs, speedup increases linearly until
the number of worker threads equals the number of cores.  After that
the increase continues at a slower rate.  For wide networks on Xeon
Phi\texttrademark, speedup increases linearly until the number of
worker threads equals the number of cores, then more slowly until
double that number, and then even slower until the number of hardware
threads.  The maximal achieved speedups for networks of different
widths are shown in Figs.~\ref{fig:2dspeedups} and
~\ref{fig:3dspeedups}.

\section{CPU vs. GPU}
While the preceding results show that ZNN can efficiently utilize
CPUs, it is also important to know how the resulting performance
compares to GPU implementations of ConvNet learning.  Therefore, we
benchmarked ZNN against Caffe \cite{jia2014caffe} and Theano
\cite{bergstra2010theano}, two popular GPU implementations.
Comparison can be tricky because CPU and GPU implementations by
definition cannot be run on the same hardware.

We chose to run Caffe and Theano on a Titan X GPU, and ZNN on an $18$
core Amazon EC2 instance (c4.8xlarge).  We chose this particular
comparison, because the alternatives seemed unfair.  For example, we
could have run ZNN on specialized hardware with more CPU cores than
the EC2 instance.  This comparison seemed unfair because the
specialized hardware would have been much more costly than Titan X and
less accessible than Amazon EC2.  Also, we could have used GPU
instances from Amazon EC2, but these are currently much slower than
Titan X ($3\times$ or more on our benchmarks) and have half the
onboard RAM.

For Caffe, both default and cuDNN\cite{chetlur2014cudnn}
implementations were used. For 3D ConvNets we only used Theano, as the
official release of Caffe still does not support 3D ConvNets.  Our
Caffe and Theano code is publicly available in the ZNN repository.

ZNN used FFT convolution for both 2D and 3D, as this was found to be
optimal by the auto-tuning capability of ZNN.  Caffe and Theano used
direct convolution.

\begin{figure}
  \centering
  \includegraphics[width=0.50\textwidth]{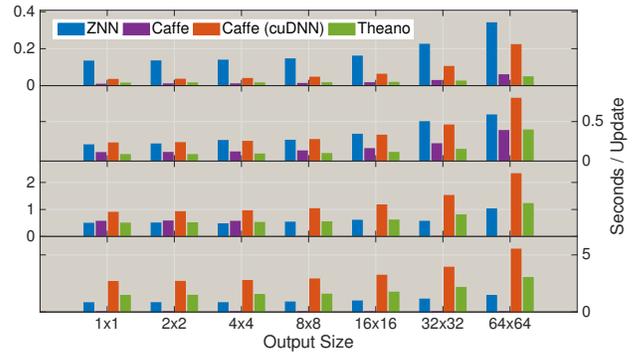}
  \caption{Comparison of ZNN, Caffe (with and without cuDNN) and
    Theano for 2D ConvNets.  The charts from the top down have kernel
    sizes of $10^2$, $20^2$, $30^2$ and $40^2$ respectively. Where
    Caffe data is missing, it means that Caffe could not handle
    networks of the given size.}
  \label{fig:vsgpu2d}
\end{figure}

Our ConvNets contained 6 fully-connected convolutional (C) layers,
each followed by a rectified linear transfer function layer (T), and
two max-pooling (P) layers, either $2 \times 2$ or $2 \times 2 \times
2$. The sequence of the layer types was CTPCTPCTCTCTCT.  All networks
had width $40$, while the sizes of the kernels and the output patch
varied.

All benchmark times were for ``sparse training,'' meaning that the
ConvNet is used to produce predictions for pixels in the output patch
that form a lattice with period 4 in every dimension.  The loss of
predicted output pixels is due to the two layers of max-pooling.

As noted before, ZNN can also perform ``dense training,'' meaning that
the ConvNet is used to produce predictions for every pixel in the
output patch by applying the ConvNet to a window that slides across
every ``valid'' location in the input patch.  Requiring Caffe or
Theano to perform dense training could have been accomplished by
computing $16$ sparse outputs in 2D and $64$ in 3D to assemble a dense
output.  This method is very inefficient and would have been no
contest with ZNN.

\subsection{Speed}
The comparison of 2D ConvNets is shown in Fig.~\ref{fig:vsgpu2d}. ZNN
is faster than Caffe and Theano for sufficiently large kernels
($30\times 30$ or larger).  This makes sense because FFT convolution
(ZNN) is more efficient than direct convolution (Caffe and Theano) for
sufficiently large kernels.

Such large kernels are not generally used in practice, so ZNN may not
be competitive with GPU implementations for 2D networks.  On the other
hand, ZNN opens up the possibility of efficiently training networks
with large kernels, and these might find some practical application in
the future.

\begin{figure}
  \centering
  \includegraphics[width=0.4\textwidth]{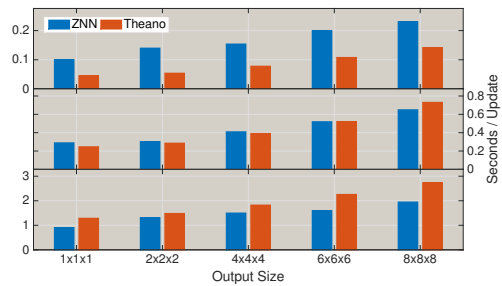}
  \caption{Comparison of ZNN and Theano for 3D ConvNets.  The charts
    from the top down have kernel sizes of $3^3$, $5^3$ and $7^3$.}
  \label{fig:vsgpu3d}
\end{figure}

The comparison of 3D ConvNets is shown in Fig.~\ref{fig:vsgpu3d}.  ZNN
is comparable to Theano even for modest kernel sizes of $5 \times 5
\times 5$ and outperforms Theano for kernel sizes of $7 \times 7
\times 7$ and greater.  Such kernel sizes are currently relevant for
practical applications~\cite{helmstaedter2013connectomic}.  Again the
benchmark makes sense, because we expect the crossover point for
complexity of FFT vs. direct convolution to occur for smaller (linear)
kernel sizes in 3D.

\subsection{Memory}
Working memory is another computational resource that is important for
training ConvNets.  Given the limited amount of onboard GPU memory, we
were unable to use Theano to train 3D networks with kernel sizes
larger than $7 \times 7 \times 7$.  We were also unable to use Caffe
to train many 2D networks (see missing bars in
Fig.~\ref{fig:vsgpu2d}).


ZNN enables training of larger networks mostly because a typical CPU
system has much more RAM than even a top GPU.  Titan X, for example,
has just 12 GB of onboard RAM.  Additionally, ZNN can achieve even
higher speed by using extra RAM space, as in the case of FFT
memoization.  When using FFT-based convolutions, with the memoization
disabled, ZNN is more efficient in its usage of RAM than the proposed
GPU methods.  The memory overhead of the methods proposed
in~\cite{mathieu-iclr-14,vasilache2014fast} could be very high as it
is proportional to the number of kernels in a layer.  In contrast
ZNN's memory overhead is proportional to the number of workers.



\section{Implementation Details}
ZNN is implemented in C++ and is publicly available under the GPL2
license (\emph{https://github.com/zlateski/znn-release}).  It can
use either fftw or intel MKL for FFTs and either provided code or
intel MKL libraries for direct convolution.  Using fftw instead of MKL
yields same scalability but lower absolute performances due to the
differences in single thread performances of the two libraries.  The
repository also provides alternative scheduling strategies such as
simple FIFO or LIFO as well as some more complex ones based on work
stealing~\cite{blumofe1999scheduling}.  The alternative scheduling
strategies achieve noticeably lower scalability than the one proposed
in the paper for most networks.  However, some very specific networks
might benefit from alternative scheduling algorithms.  Future work can
include automatic detection of the best scheduling strategy.


\section{Conclusions}

ZNN achieves high performances by efficiently utilizing the available
CPUs.  We expect an increase in the number of cores per chip (or Xeon
Phi\texttrademark card) in the future, making ZNN even more practical.
In fact, we have already used ZNN to achieve state of the art results
in boundary detection~\cite{lee2015recursive} and computation of
dendritic arbor densities~\cite{sumbul2014automated}.

Having a large amount of RAM available to the CPU, ZNN can efficiently
train very large ConvNets with large kernels. ZNN allows for easy
extensions and can efficiently train a ConvNet with an arbitrary
topology, allowing for new research.

Unlike the ZNN's task parallelization model, the current GPU
implementations employ SIMD parallelism to perform computation on one
whole layer at a time, thus limiting the network structure.  Mainly,
the computation is parallelized such that a single thread computes the
value of a single voxel of an output image.  Libraries like cuDNN
provide optimized primitives for fully connected convolutional layers
by reducing all the required convolutions in the layer to a matrix
multiplication, which is then parallelized on the GPU.

Extending the functionality requires the user to provide a
parallelized implementation of the new layer type, which typically
requires great knowledge of GPU programming, and might take a long
time.  Contrary to that, ZNN's task parallelism allows for easy
extensions by simply providing serial functions for the forward and
backward pass, as well as the gradient computation, if required.
ZNN's repository contains some sample extensions providing
functionality of \emph{dropout}~\cite{srivastava2014dropout} and
\emph{multi-scale}~\cite{long2015fully,sermanet2011traffic} networks.



\begin{thebibliography}{10}

\bibitem{jia2014caffe}
Y.~Jia, E.~Shelhamer, J.~Donahue, S.~Karayev, J.~Long, R.~Girshick,
  S.~Guadarrama, and T.~Darrell, ``Caffe: Convolutional architecture for fast
  feature embedding,'' in {\em Proceedings of the ACM International Conference
  on Multimedia}, pp.~675--678, ACM, 2014.

\bibitem{collobert2011torch7}
R.~Collobert, K.~Kavukcuoglu, and C.~Farabet, ``Torch7: A matlab-like
  environment for machine learning,'' in {\em BigLearn, NIPS Workshop},
  no.~EPFL-CONF-192376, 2011.

\bibitem{bergstra2010theano}
J.~Bergstra, O.~Breuleux, F.~Bastien, P.~Lamblin, R.~Pascanu, G.~Desjardins,
  J.~Turian, D.~Warde-Farley, and Y.~Bengio, ``Theano: a cpu and gpu math
  expression compiler,'' in {\em Proceedings of the Python for scientific
  computing conference (SciPy)}, vol.~4, p.~3, Austin, TX, 2010.

\bibitem{dean2012large}
J.~Dean, G.~Corrado, R.~Monga, K.~Chen, M.~Devin, M.~Mao, A.~Senior, P.~Tucker,
  K.~Yang, Q.~V. Le, {\em et~al.}, ``Large scale distributed deep networks,''
  in {\em Advances in Neural Information Processing Systems}, pp.~1223--1231,
  2012.

\bibitem{mathieu-iclr-14}
M.~Mathieu, M.~Henaff, and Y.~LeCun, ``Fast training of convolutional networks
  through ffts,'' in {\em International Conference on Learning Representations
  (ICLR2014)}, CBLS, April 2014.

\bibitem{vasilache2014fast}
N.~Vasilache, J.~Johnson, M.~Mathieu, S.~Chintala, S.~Piantino, and Y.~LeCun,
  ``Fast convolutional nets with fbfft: A gpu performance evaluation,'' {\em
  arXiv preprint arXiv:1412.7580}, 2014.

\bibitem{masci2013fast}
J.~Masci, A.~Giusti, D.~Ciresan, G.~Fricout, and J.~Schmidhuber, ``A fast
  learning algorithm for image segmentation with max-pooling convolutional
  networks,'' in {\em Image Processing (ICIP), 2013 20th IEEE International
  Conference on}, pp.~2713--2717, IEEE, 2013.

\bibitem{giusti2013fast}
A.~Giusti, D.~C. Cire{\c{s}}an, J.~Masci, L.~M. Gambardella, and
  J.~Schmidhuber, ``Fast image scanning with deep max-pooling convolutional
  neural networks,'' {\em arXiv preprint arXiv:1302.1700}, 2013.

\bibitem{sermanet2013overfeat}
P.~Sermanet, D.~Eigen, X.~Zhang, M.~Mathieu, R.~Fergus, and Y.~LeCun,
  ``Overfeat: Integrated recognition, localization and detection using
  convolutional networks,'' {\em arXiv preprint arXiv:1312.6229}, 2013.

\bibitem{viebke2015potential}
A.~Viebke and S.~Pllana, ``The potential of the intel xeon phi for supervised
  deep learning,'' {\em arXiv preprint arXiv:1506.09067}, 2015.

\bibitem{Jin:2014:TLS:2672598.2672903}
L.~Jin, Z.~Wang, R.~Gu, C.~Yuan, and Y.~Huang, ``Training large scale deep
  neural networks on the intel xeon phi many-core coprocessor,'' in {\em
  Proceedings of the 2014 IEEE International Parallel \& Distributed Processing
  Symposium Workshops}, IPDPSW '14, (Washington, DC, USA), pp.~1622--1630, IEEE
  Computer Society, 2014.

\bibitem{krizhevsky2012imagenet}
A.~Krizhevsky, I.~Sutskever, and G.~E. Hinton, ``Imagenet classification with
  deep convolutional neural networks,'' in {\em Advances in neural information
  processing systems}, pp.~1097--1105, 2012.

\bibitem{ciresan2012deep}
D.~Ciresan, A.~Giusti, L.~M. Gambardella, and J.~Schmidhuber, ``Deep neural
  networks segment neuronal membranes in electron microscopy images,'' in {\em
  Advances in neural information processing systems}, pp.~2843--2851, 2012.

\bibitem{long2015fully}
J.~Long, E.~Shelhamer, and T.~Darrell, ``Fully convolutional networks for
  semantic segmentation,'' in {\em The IEEE Conference on Computer Vision and
  Pattern Recognition (CVPR)}, June 2015.

\bibitem{kanazawa2014locally}
A.~Kanazawa, A.~Sharma, and D.~W. Jacobs, ``Locally scale-invariant
  convolutional neural networks,'' {\em arXiv preprint arXiv:1412.5104}, 2014.

\bibitem{sermanet2011traffic}
P.~Sermanet and Y.~LeCun, ``Traffic sign recognition with multi-scale
  convolutional networks,'' in {\em Neural Networks (IJCNN), The 2011
  International Joint Conference on}, pp.~2809--2813, IEEE, 2011.

\bibitem{bretttheorem}
J.~Gustafson, ``Brent’s theorem,'' in {\em Encyclopedia of Parallel
  Computing} (D.~Padua, ed.), pp.~182--185, Springer US, 2011.

\bibitem{michael1996simple}
M.~M. Michael and M.~L. Scott, ``Simple, fast, and practical non-blocking and
  blocking concurrent queue algorithms,'' in {\em Proceedings of the fifteenth
  annual ACM symposium on Principles of distributed computing}, pp.~267--275,
  ACM, 1996.

\bibitem{boost_lockfree}
T.~Blechmann, ``Boost lockfree library.'' http://www.boost.org/libs/lockfree/,
  2008.

\bibitem{chetlur2014cudnn}
S.~Chetlur, C.~Woolley, P.~Vandermersch, J.~Cohen, J.~Tran, B.~Catanzaro, and
  E.~Shelhamer, ``cudnn: Efficient primitives for deep learning,'' {\em arXiv
  preprint arXiv:1410.0759}, 2014.

\bibitem{helmstaedter2013connectomic}
M.~Helmstaedter, K.~L. Briggman, S.~C. Turaga, V.~Jain, H.~S. Seung, and
  W.~Denk, ``Connectomic reconstruction of the inner plexiform layer in the
  mouse retina,'' {\em Nature}, vol.~500, no.~7461, pp.~168--174, 2013.

\bibitem{blumofe1999scheduling}
R.~D. Blumofe and C.~E. Leiserson, ``Scheduling multithreaded computations by
  work stealing,'' {\em Journal of the ACM (JACM)}, vol.~46, no.~5,
  pp.~720--748, 1999.

\bibitem{lee2015recursive}
K.~Lee, A.~Zlateski, A.~Vishwanathan, and H.~S. Seung, ``Recursive training of
  2d-3d convolutional networks for neuronal boundary detection,'' {\em arXiv
  preprint arXiv:1508.04843}, 2015.

\bibitem{sumbul2014automated}
U.~S{\"u}mb{\"u}l, A.~Zlateski, A.~Vishwanathan, R.~H. Masland, and H.~S.
  Seung, ``Automated computation of arbor densities: a step toward identifying
  neuronal cell types,'' {\em Frontiers in neuroanatomy}, vol.~8, 2014.

\bibitem{srivastava2014dropout}
N.~Srivastava, G.~Hinton, A.~Krizhevsky, I.~Sutskever, and R.~Salakhutdinov,
  ``Dropout: A simple way to prevent neural networks from overfitting,'' {\em
  The Journal of Machine Learning Research}, vol.~15, no.~1, pp.~1929--1958,
  2014.

\end{thebibliography}
\end{document}